\documentclass[preprint]{elsarticle}

\usepackage[utf8]{inputenc}
\usepackage{amssymb, amsthm, mathtools}
\usepackage{bm}
\usepackage{booktabs}
\usepackage{float}

\usepackage{caption}
\usepackage{subcaption}
\usepackage{algorithm}
\usepackage{algpseudocode}
\usepackage{multirow}

\usepackage{url}
\usepackage{xcolor}

\DeclareMathOperator*{\argmin}{arg\,min}

\theoremstyle{plain}
\newtheorem{theorem}{Theorem}
\newtheorem{proposition}{Proposition}

\journal{Neurocomputing}

\begin{document}
\begin{frontmatter}
    \title{Denoising Score Matching with Random Fourier Features}
    %\author{Tsimboy Olga, Yermek Kapushev, Evgeny Burnaev, Ivan Oseledets}
    %% Group authors per affiliation:
    \author[skoltech]{Tsimboy Olga}
    \author[skoltech]{Yermek Kapushev\corref{correspondingauthor_second}}
    \ead{y.kapushev@skoltech.ru}
    \author[skoltech]{Evgeny Burnaev}
    \author[skoltech]{Ivan Oseledets}
    \address[skoltech]{Skolkovo Institute of Science and Technology,
                       Bolshoy Boulevard 30, bld. 1,
                       Moscow, Russia}

    \cortext[correspondingauthor_second]{Authors contributed equally}

    \begin{abstract}
        The density estimation is one of the core problems in statistics.
        Despite this, existing techniques like maximum likelihood estimation are
        computationally inefficient due to the intractability of the normalizing constant.
        For this reason an interest to score matching has increased being independent on the normalizing constant.
        However, such estimator is consistent only for distributions with the full space support.
        One of the approaches to make it consistent is to add noise to the input data which is called Denoising Score Matching.

        In this work we derive analytical expression for the Denoising Score matching
        using the Kernel Exponential Family as a model distribution.
        The usage of the kernel exponential family is motivated by the richness of this class of densities.
        To tackle the computational complexity we use Random Fourier Features based approximation of the kernel function.
        The analytical expression allows to drop additional regularization terms based on the higher-order derivatives
        as they are already implicitly included.
        Moreover, the obtained expression explicitly depends on the noise variance, so the validation loss can be
        straightforwardly used to tune the noise level.
        Along with benchmark experiments, the model was tested on various synthetic distributions
        to study the behaviour of the model in different cases.
        The empirical study shows comparable quality to the competing approaches, while the proposed method being computationally faster.
        The latter one enables scaling up to complex high-dimensional data.
    \end{abstract}

    \begin{keyword}
    \end{keyword}
\end{frontmatter}

% \linenumbers

\section{Introduction}

One of the core problems in statistics is a density estimation.
The most well-known approach is the Maximum Likelihood Estimation (MLE).
However, MLE and all other approaches based on MLE require normalizing constant to be known
or computed efficiently, which is not the case in many real world problems.
The intractability of the normalizing constant makes the approach infeasible.
In contrast, an unsupervised score matching estimator~\cite{Hyvarinen2005} based on Fisher divergence minimization, does not depend on the normalizing constant.
The resulting estimate is proved to be asymptotically normal and consistent in the case when
data and model distributions supports coincide.
There numerous developments of the
 idea~\cite{hyvarinen2007some, Lyu2012, Gutmann2012bregman, Kanti2016, dai2018kernel}.

Another important part of the density estimation is the class of models to search the solution in.
A special interest is paid here to an exponential family of distributions which leads to the
closed-form solution~\cite{hyvarinen2007some, forbes2015linear, Lin2016, Shiqing2018, monti2018}.
A generalization of the finite-dimensional exponential families is a kernel exponential family
(KEF).
In this case the natural parameter is treated a function from some
Reproducing Kernel Hilbert Space (RKHS).
It can be seen as infinite-dimensional generalization of the exponential family.
The KEF contains all well-known exponential family densities such as Exponential, Gaussian,
Gamma, etc.
In addition, RKHS reveals a sufficiently rich class of
estimators with convergence guarantees w.r.t. different
metrics~\cite{Gretton2013, Gretton2015}.
The main disadvantage is the computational cost for the sample matrix inversion making this
method inapplicable even for a moderate amount of the training data.

To approach the computational complexity issue~\cite{sutherland2017efficient, GrettonDeep} propose to use Nystr{\"o}m-type approximation of the kernel function.
Alternatively, Random Fourier Features (RFF)~\cite{RFF} embedding can be considered.
Employing special structure on the RFF~\cite{choromanski2017unreasonable,munkhoeva2018quadrature}
we come up with a faster model than Nystr{\"o}-type approximation.
There are a lot of papers studying the convergence of the RFF models for the regression problem.
The optimal learning rate with $\mathcal{O}(\sqrt{n}\log n)$ features is the same as for the
full kernel \cite{aless2016generalization}
which gives substantial speed up.
The theoretical properties of using RFF for score matching is less studied,
though there are some general theoretical results on RFF and
higher order kernel derivatives~\cite{Orlicz,OperatorValuedKernels}.

Naive approach to score matching with RFF suffer from several issues.
The first one is an oscillating behaviour in the tails of the distribution~\cite{Gretton2015}.
The second problem is poor convergence in the case of disjoint support
(consistency could not be guaranteed) or in the areas
where density value is close to zero~\cite{GrettonDeep}.
Inconsistency explains low approximation accuracy in regions of almost zero density.

It was shown that convolution with small Gaussian noise
(which is equivalent to the noisy data perturbation) improves learning behaviour and approximation
quality, e.g.~\cite{song2019generative,GANinstability, NoisyFGAN}.
It makes the support of both densities (distribution of the data and the model distribution)
the same and allows to overcome the aforementioned issue.
For most of the models the convolution cannot be calculated analytically,
so authors usually stick to the second-order Taylor series expansion
\cite{NIPS2010_4060, Reehorst_2019, NoisyFGAN} which results in a special
regularization term in the loss function.
It turns out that the noise level is an important parameter.
With large noise level we have better convergence but lower accuracy.
With smaller noise level the convergence is less stable, but the
solution is more accurate.
This means, that tuning of noise level is required.
Recently, it was proposed to use several noise levels optimizing
cumulative objective~\cite{song2019generative}.
% In this work we show, that for kernel exponential family we can find
% the optimal noise parameters by a simple gradient-based approach.

\paragraph{Contribution}
In thiw work we introduce method to estimate unknown distribution using
denoising score matching combined with RFF.
To tackle the convergence issues we convolve the loss function with symmetric noise analytically.
It allows to avoid additional regularization terms as they are embedded into the loss function naturally.
The derived expression of the loss function explicitly contains the noise parameters
that allows us to use simple gradient-based approaches to tune these parameters.
In the experimental section we demonstrate the performance of our approach both in terms of
accuracy and training time.
While the quality is comparable to Nystr{\"o}m-type approximations, the training speed is much faster.

The paper is organized as follows.
In Section~\ref{sec:background} we give background information
that is used to construct the final model:
score-matching and its RKHS form, learning using random features.
Section~\ref{chap:body} provides results on the necessary condition of denoising score matching
and its RFF approximation.
Numerical experiments are presented in Section~\ref{chap:experiments}.
Finally, Section~\ref{chap:Conclusion} concludes the results of conducted research.
All additional materials are presented in
appendices~\ref{chap:Technical},~\ref{sec:B},~\ref{sec:C}.

\section{Background}
\label{sec:background}
\subsection{Score matching}
Let ${\cal D} = \{{\bf x}_a\}_{a = 1}^d, {\bf x}_a \in \mathbb{R}^d$ be a set of observations
drawn from an unknown distribution with a probability density function $p_0({\bf x})$.
Let $p({\bf x}, {\bm \theta})$ be a model density parameterized by
${\bm \theta} \in \Theta \subset \mathbb{R}^m$.
The task is to find such ${\bm \theta}^*$ that the model density is close to the real one:
$p({\bf x}, {\bm \theta}^*) \approx p_0({\bf x})$.
In score matching approach we minimize the Fisher divergence:
\begin{equation}
    J(p_0 \| p_{\bm{\theta}}) = \frac12 \int p_0({\bf x})\|\nabla \log p({\bf x}, {\bm \theta}) - \nabla \log p_0({\bf x})\|_2^2 d\bm{x}.
    \label{eq:fisher}
\end{equation}
Under sufficiently weak regularity conditions (see \cite{Hyvarinen2005})
the minimization of the Fisher divergence is equivalent to minimization of
\begin{equation}
    J(p_0 \| p_{\bm{\theta}}) \sim \mathbb{E}_{p_0}\left[\Delta \log p({\bf x}, {\bm \theta}) + \frac{1}{2}\|\nabla \log p({\bf x}, {\bm \theta})\|^2\right]
    \label{eq:fisher_sm}.
\end{equation}
Note, that the normalizing constant does not depend on ${\bf x}$,
therefore, $p({\bf x}, {\bm \theta})$ in \eqref{eq:fisher_sm}
could be replaced with unnormalized one
$\Tilde{p}({\bf x}, {\bm \theta}) = p({\bf x}, {\bm \theta}) Z({\bm \theta})$.
In an abuse of notation from now on we will use
$p({\bf x}, {\bm \theta})$ to denote the unnormalized density if it not stated explicitly.
% $\Tilde{p}({\bf x}, {\bm \theta})$ as .
Objective \eqref{eq:fisher_sm} now does not depend on unknown density $p_0$ and provides an
opportunity to estimate $p_0$ up to the normalizing constant using only samples drawn from $p_0$:
\begin{equation}
    \hat{J}(p_0 \| p_{\bm{\theta}}) = \frac{1}{n}\sum_{a = 1}^n\left[\Delta \log p({\bf x}_a, {\bm \theta}) + \frac{1}{2}\|\nabla \log p({\bf x}_a, {\bm \theta})\|^2\right] \to \min_{{\bm \theta}}.
    \label{eq:fisher_sm_sample}
\end{equation}
This loss suffers from several issues.
Firstly, the expression \eqref{eq:fisher} assumes that model and data distributions
have the same support.
However, in real world the real distribution lies on a low-dimensional manifold embedded
in $\mathbb{R}^d$ \cite{song2019generative},
while support of the model density is usually the whole space.
Secondly, score matching convergence is guaranteed only in the case of
${\rm supp\;} p_0 = \mathbb{R}^d$ (see \cite{Hyvarinen2005}).

To tackle the issue we use Denoising Score Matching (DSM) \cite{Denoising}.
In this approach we add noise to the data.
The score matching loss in this case is given by
\begin{equation}
    \label{eq:denoising_sm}
    DSM(p_{\bm{\theta}}) = \mathbb{E}_{p_{\varepsilon}}\mathbb{E}_{p_0}\left[\Delta \log p({\bf x} + {\bm \varepsilon}, {\bm \theta}) + \frac{1}{2}\|\nabla \log p({\bf x} + {\bm \varepsilon}, {\bm \theta})\|^2\right],
\end{equation}
where $p_{\varepsilon}(\bm{x})$ is a distribution of noise.
Now both densities have the same support, so the solution converges.
The optimal model satisfies
$\nabla p_{\bm{\theta}} = \nabla \left [p_0 * p_{\varepsilon} \right ] (\bm{x})$,
where $*$ is the convolution operator.
However, $\nabla \left [p_0 * p_{\varepsilon} \right ] (\bm{x})$ is
close to the true density $\nabla p_0(\bm{x})$
only when the noise is small enough.

To estimate the loss in general case we can generate finite set of
noisy samples and use them to estimate expectation in the loss function.
Another option is to use Taylor series expansion assuming that noise level is small.
In both cases we get approximate value of the loss function.
Moreover, when we use Taylor series expansion we need to calculate
higher order derivatives of the model which can be computationally complex
(for example, in case of neural networks).
However, for the kernel exponential family the denoising score matching loss
can be computed exactly.

\subsection{Kernel exponential family}
The kernel exponential family is a set of distributions where unnormalized
probability density functions
$p_f(\bf{x})$ satisfy
$\log p_f({\bf x}) = f({\bf x}) + \log q_0({\bf x})$,
$f \in \mathcal{H}$,
$\mathcal{H}$ is some Reproducing Kernel Hilbert Space (RKHS) with kernel $k$ and
$q_0$ is some generating density.
The normalizing constant is usually not known and cannot be computed analytically.
The class of such densities is rich enough.
In fact it is dense in a set of continuous probability density functions
that decay at the same rate as $q_0$.

In a well specified case, i.e. $p_0$ belongs to the kernel exponential family with RKHS
$\mathcal{H}$, the score matching loss \eqref{eq:fisher_sm}
can be expressed as (see \cite{Gretton2013})
\begin{equation}
    \label{eq:sm_kernel}
    J(p_0 \| p_f) =
    \frac{1}{2} \langle f, Cf\rangle_{\mathcal{H}} + \langle f, \xi\rangle_{\mathcal{H}} + J(p_0 \| q_0)
\end{equation}
where
$\partial^{\alpha,\beta}_{i, j + d} k({\bf x, \bf y}) =
\frac{\partial^{\alpha + \beta}}{\partial x_i^{\alpha}\partial y_j^{\beta}}k({\bf x, \bf y})$
and
\begin{align*}
    C &= \mathbb{E}_{p_0} \left[
        \sum_{i = 1}^d \partial_i k ({\bf  x}, \cdot) \otimes \partial_i k ({\bf  x}, \cdot)
    \right], \quad C \colon \mathcal{H} \to \mathcal{H} \\
    \xi &= \mathbb{E}_{p_0} \left[
        \sum_{i = 1}^d \partial_i k ({\bf  x}, \cdot)\partial_i \log q_0({\bf x}) +
        \partial^2_i k ({\bf  x}, \cdot)
        \right] \in \mathcal{H}.
\end{align*}

Using the general representer theorem the optimal $f({\bf x})$ can be found
as a weighted sum of the kernel derivatives located at the training samples.
To find the weights we need to invert $nd \times nd$ matrix and the computational
complexity, therefore, is $O(n^3d^3)$.
While the convergence in RKHS of this estimator implies the convergence in $L^r$,
in terms of Kullback-Leibler divergence and Hellinger distance,
in the misspecified case a density estimator remains the same,
but with convergence guarantees only for Fisher divergence.

To reduce complexity the authors of \cite{sutherland2017efficient} proposed to
find solution in a span over a randomly selected subset of training samples (inducing points).
The computational cost of this approach is $O(m^3d^3)$,
where $m$ is a number of inducing points.
Additional sub-sampling over $md$ basis functions enables even more computationally efficient
approach.
In this extreme case the complexity is $O(m^2nd + m^3)$.
As in the case of full data usage, obtained estimator is consistent when $p_0$
lies in the kernel exponential family,
but the rate of convergence is slower (under assumptions presented in \cite{sutherland2017efficient}).
The misspecified case was not studied.

To obtain consistent estimator from the kernel exponential family
the authors of \cite{GrettonDeep} used denoising score matching with Taylor series expansion.
This results in an additional regularization term in the loss function
that penalizes second derivatives of the model.
The need to calculate second derivatives restricts the approach only to
relatively low-dimensional cases.

%\newpage
\subsection{Random Fourier Features}
\label{sec:inrto_rff}
In general, there are two types of approaches to scale up the kernel methods:
Nysrt\"{o}m-type approximation \cite{sutherland2017efficient} and
random features based approximations  \cite{RFF, munkhoeva2018quadrature}.
The second one is data-independent, the idea of which follow from Bochner's theorem
\cite{Rudin1962FourierAO}: any shift-invariant bounded continuous kernel
$k({\bf x}, {\bf y}) = k({\bf x - y})$
is a Fourier transform of a non-negative bounded measure $p(\bf{w})$
\[
    k({\bf x - y}) = \int p({\bf w}) e^{j\bf{w^\top(x - y)}}d\bf{w},
\]
where $k(0) = 1$.

Using Monte-Carlo to estimate the integral we obtain the Random Fourier Features approximation
of the kernel function
\begin{equation}
\label{eq:bochner_theorem}
    k({\bf x - y}) \approx \phi({\bf x})^\top \phi({\bf y}),
    \quad
    \phi({\bf x}) = \phi(\bm{W}\mathbf{x} + {\bm b}) = \sqrt{\frac{2}{M}} \cos (\bm{W}\mathbf{x} + {\bm b}),
\end{equation}
where
\begin{align*}\quad
    \bm{W} = \begin{bmatrix}
        \mathbf{w}_1^\top \\
        \cdots \\
        \mathbf{w}_M^\top
    \end{bmatrix},
    \quad
    \bm{b} = \begin{bmatrix}
        b_1 \\
        \cdots \\
        b_M
    \end{bmatrix},
    \quad {\bf w}_j \sim p({\bf w}), \quad b_j \sim U[0, 2 \pi].
\end{align*}
% This choose of RFF representation is motivated only by further simplicity of derivations that
% could be easily extended to the explicit feature map with lower estimation error
% \cite{sutherl2015errorRFF}, another variant of error reduction is to use am importance
% sampling which implies to weighted RFF \cite{avron2018random}.
% Kernel derivatives approximation is defined in the same form as~\eqref{eq:bochner_theorem},
% with convergence bound for a variety of kernels \cite{Orlicz}.
The same idea can be applied to the kernel derivatives
\begin{equation}
    \partial^{{\bf p}, {\bf q}} k({\bf x - y}) = \int p({\bf w}) \partial^{{\bf p}}\left[e^{j\bf{w^\top x}}\right]\partial^{{\bf q}}\left[e^{-j\bf{w^\top y}}\right]d\bf{w}
\end{equation}
where ${\bf p}, {\bf q} \in \mathbb{R}^d$ denote multi-indices,
$\partial^{\bf{p}} f =
\frac{\partial^{|\bf{p}_1 + \bf{p}_2 + \cdots + \bf{p}_d |} }{\partial x_1^{p_1} \ldots
\partial x_d^{p_d}} f$
and $\bf{p}$ and $\bf{q}$ act on the first and the second arguments of
the kernel correspondingly.

RFF sampling could be improved using the special structure on weights ${\bf w}_j$ and better
expectation approximation in \eqref{eq:bochner_theorem}
\cite{yu2016orthogonal, munkhoeva2018quadrature, choromanski2017unreasonable} which makes RFF
generation faster than Nystr\"om approach.
Nevertheless, the theoretical properties are well studied only for the kernel ridge regression
\cite{aless2016generalization, li2018unified}.
Also random features approach could be extended to more general class of kernels
that admits representation
$k(\bf{x}, \bf{y}) = \int \phi(\bf{w}, \bf{x})^\top \phi(\bf{w}, \bf{y}) d\pi(\bf{w})$
for some feature map $\phi(\bf{w}, \cdot)$ and measure $\pi(\bf{w})$.
One of the most well-known kernels of such type is the Arc-Cosine kernel
\cite{Youngmin}
\begin{equation}
\label{eq:arccos_ker}
    k({\bf x}, {\bf y}) = \int \phi({\bf x}, {\bf w})\phi({\bf y}, {\bf w}) d\pi({\bf w}) =
    \frac{1}{\pi}\|{\bf x}\|^p\|{\bf y}\|^pJ_p(\theta({\bf x}, {\bf y})),
\end{equation}
\[
    J_p(\theta) = (-1)^p \left(\sin \theta\right)^{2p + 1} \left (
            \frac{1}{\sin \theta} \frac{\partial}{\partial \theta}
        \right )^p \left (
            \frac{\pi - \theta}{\sin \theta}
        \right ),
\]
\begin{equation}
    \label{eq:arccos_features}
    \phi({\bf x}, {\bf w}) = ({\bf w}^{\top}{\bf x})^p{\bf 1}({\bf w}^{\top}{\bf x}),
    \quad {\bf w},{\bf x} \in \mathbb{R}^d,~p \geq 0,
\end{equation}

where $\pi({\bf w}) = {\mathcal N}({\bf 0}, {\bf I}_d)$
and $\theta({\bf x}, {\bf y})$ is an angle between vectors ${\bf x}$ and ${\bf y}$.

% It should be noticed that in \cite{williams2001using, GrettonDeep} for kernel exponential family there was used not the kernel, but model Nystr\"om approximation: basis functions sub-sampling strategies were proposed to reduce computational complexity.
\section{Kernel Denoising Score Matching}
\label{chap:body}
This section provides the optimal solution for the Kernel Denoising Score Matching,
its RFF approximation and some error bounds of the resulting model.
Here we assume that the noise distribution is symmetric, i.e.
$p_{\varepsilon}(\bf{x}) = p_{\varepsilon}(-\bf{x})$.

\subsection{Denoising Score Matching in RKHS}
We start from rewriting the expression for the Denoising Score Matching objective
\eqref{eq:denoising_sm}
and follow the same logic in derivation as in paper \cite{Gretton2013}
with the difference that our objective function is the convolution of the usual
score matching objective with noise distribution.

Let $V: \mathbb{R}^m \to \mathbb{R}$ be a convex and differentiable function.
Assume that the objective function takes the form
\[
    J(f) = V(\langle \phi_1, f\rangle_\mathcal{H},
             \langle \phi_2, f\rangle_\mathcal{H}, \ldots,
             \langle \phi_m, f\rangle_\mathcal{H}) + \frac{\lambda}{2}\|f\|^2_\mathcal{H},
\]
for any set $\{\phi_i(\cdot)\}_{i = 1}^m$, $\phi_i \in \mathcal{H}$.

For our case we define the set of functions $\{\phi_i(\cdot)\}$ as follows
\begin{align*}
    &\phi_{(a - 1)d + i}(\cdot) = \partial_i k({\bf x}_a + {\bf y}, \cdot),\\
    &\phi_{nd + 1}(\cdot) = \frac{1}{n}\sum\limits_{a, i = 1}^{n, d} \partial_i^2 k({\bf x}_a + {\bf y}, \cdot) + \partial_i k({\bf x}_a + {\bf y}, \cdot)\partial_i \log q_0({\bf x}_a + {\bf y}),
\end{align*}
and for simplicity let us denote it as $\{\phi_i(y, \cdot)\}_{i = 1}^m$, $m = nd + 1$.
Now let us define a linear operator
$A({\bf y}): \mathcal{H} \to \mathbb{R}^m$, $f \to \{\langle\phi_i({\bf y}, \cdot),
f\rangle_\mathcal{H}\}_{i = 1}^m$.
Then the objective \eqref{eq:denoising_sm} can be written as
\begin{equation}
    \label{eq:denoising_objective_rkhs}
    f^* = \argmin_{f \in \mathcal{H}} \int p_{\varepsilon}({\bf y})V(A({\bf y})f)d{\bf y}
    + \frac{\lambda}{2}\|f\|^2,
\end{equation}
with $V(\theta_1, \ldots, \theta_{nd + 1}) =
\frac{1}{2n}\sum\limits_{a = 1}^n\sum\limits_{i = 1}^d\theta^2_{(a - 1)d + i} +
\theta_{nd + 1}$.

Using the first order optimality condition we can see that the solution takes the form
\[
    f = \int p_{\varepsilon}({\bf y})A^*({\bf y}){\bm \alpha}({\bf y})d {\bf y},
    \quad
    {\bm \alpha}({\bf y}) = -\frac{1}{\lambda}\nabla V(A({\bf y})f),
\]
where $A^*({\bf y}): \mathbb{R}^m \to \mathcal{H}$ is an adjoint to $A({\bf y})$.
% and
% $A^*({\bf y}){\bm \alpha} = \sum_{i = 1}^m \alpha_i\phi_i({\bf y}, \cdot)$, denoting
% \[
%     {\bm \alpha}({\bf y}) = -\frac{1}{\lambda}\nabla V(A({\bf y})f),\quad f = \int p_{\varepsilon}({\bf y})A^*({\bf y}){\bm \alpha}({\bf y})d {\bf y}
% \]
% first order optimality condition could be written as an integral equation on ${\bm \alpha}({\bf y})$:
% \begin{equation}
%     {\bm \alpha}({\bf y}) = -\frac{1}{\lambda}\nabla V\left(\int p_{\varepsilon}({\bf z})A({\bf y})A^*({\bf z})\alpha({\bf z})d{\bf z}\right)
% \end{equation}
% where $A({\bf y})A^*({\bf z})\alpha({\bf z}) = \sum\limits_{i = 1}^m\alpha_i({\bf z})\{\langle \phi_j({\bf y}, \cdot),  \phi_i({\bf z}, \cdot)\rangle\}_{j = 1}^m = {\bm K}({\bf y}, {\bf z}){\bm \alpha}({\bf z})$.
Now we are ready to formulate the proposition.
\begin{proposition}
    The solution to \eqref{eq:denoising_objective_rkhs} has the following form
    \[
        f^* = B \left [
            -\frac{1}{n\lambda}C(\bm{\beta}^*) + \frac{1}{n\lambda^2}b
        \right ],
    \]
    where
    $\hat{A}({\bf y}) \colon \mathcal{H} \to \mathbb{R}^{m - 1}$,
    $\left ( \hat{A}({\bf y})f \right )_i = \left( A(\bf{y})f \right )_i, i=1, \ldots, m-1$,
    \newline
    $B = \int p_{\varepsilon}({\bf y})\hat{A}^*({\bf y})\hat{A}({\bf y}) d{\bf y}$,
    $b = \int p_{\varepsilon}({\bf y})\phi_m({\bf y}, \cdot) d {\bf y}$,
    $C(\bm{\beta}) = \int p_{\varepsilon}({\bf x})\hat{A}({\bf y})\bm{\beta}({\bf y})d{\bf y}$
    and ${\bm\beta}^*({\bf y})$ is the solution to
    \begin{equation}
        \bm{\beta}({\bf y}) = -\frac{1}{n\lambda}
        \int p_{\varepsilon}({\bf z}) \hat{A}({\bf y})\hat{A}({\bf z})^*\bm{\beta}({\bf z})
        d{\bf z}
        + \frac{1}{n\lambda^2}
        \int p_{\varepsilon}({\bf z}) \hat{A}({ \bf y}) \phi_m({\bf z}, \cdot)d{\bf z}.
    \label{eq:beta_integal}
    \end{equation}
\end{proposition}
See the details on derivation in \ref{sec:exact_solution}.

The optimal model requires solution of operator equation and in general
this is a difficult task.
In order to avoid this, let us consider a Monte-Carlo approximation
of~\eqref{eq:beta_integal}.
Suppose we sampled $K$ noise vectors $\{{\bf z}_k\}_{k = 1}^K$,
${\bf z}_k \sim p_{\varepsilon}$.
In this case the approximation to the optimal ${\bm \beta}^*$ can be found
by solving the system of equations
\begin{equation}
    \label{eq:beta_finite_sample}
    {\bm \beta}_K({\bf y}) = -\frac{1}{nK\lambda}
    \sum\limits_{k = 1}^K
    \hat{A}({\bf y})\hat{A}({\bf z}_k)^*{\bm \beta}_K({\bf z}_k)
    + \frac{1}{n\lambda^2}\int p_{\varepsilon}({\bf z}) \hat{A}({\bf y}) \phi_m({\bf z}, \cdot)d{\bf z}.
\end{equation}
The obtained result can then be used to derive an approximation of $f^*$,
but the computational complexity is $O(n^3d^3K^3 + n^2d^2K)$.
Moreover, the convolution in the second term of \eqref{eq:beta_finite_sample} could be directly computed
only for a limited set of kernels, e.g. Radial Basis Function kernel (RBF).

In order to improve the computational complexity we employ RFF approach
to the kernel function approximation.

\subsection{RFF for Denoising Score Matching}
For the RFF \eqref{eq:bochner_theorem}
we introduce the following matrix of RFF derivatives $\partial {\bm \Phi}_y$
corrupted by noise $\bf{y}$.
The $((a - 1)d + i)$-th row of matrix $\partial {\bm \Phi}_y$ is given by
$[\partial \bm{\Phi}_y]_{(a - 1)d + i} = \partial_i \bm{\phi}^\top(\bm{W}({\bf x}_a + \bf{y}) + \bm{b})$,
where $\partial_i \bm{\phi}^\top(\bm{W}({\bf x}_a + \bf{y}) + \bm{b})$
is an element-wise partial derivative of the feature vector at point ${\bf x}_a$.
Similarly, for the second derivatives we have
$[\partial^2 \bm{\Phi}_y]_{(a - 1)d + i} =
\partial^2_i \bm{\phi}^\top(\bm{W}({\bf x}_a + \bf{y}) + \bm{b})$.
The finite sample solution to \eqref{eq:beta_finite_sample} is given by
\[
    f_K
    = \frac{1}{n\lambda^2}{\bm \phi}(\cdot)^{\top}{\bm H}
    \left[-\frac{1}{K}\left(\frac{1}{K}\partial  {\bm \Phi}_K^{\top}\partial{\bm \Phi}_K + n\lambda {\bf I}\right)^{-1}\partial {\bm \Phi}_K^{\top}\partial{\bm \Phi}_K \odot {\bm h} + {\bm h}\right]
    -\frac{1}{\lambda} {\bm \phi}(\cdot)^{\top}{\bm h}.
\]
Here operator $\odot$ denotes the Hadamard product.
Let us denote
\begin{equation}
    \label{eq:h}
    {\bm H} = \int p_{\varepsilon}({\bf y})\partial{\bm \Phi}_y^{\top}\partial{\bm \Phi}_y
    d{\bf y},
    \quad
    {\bm h} = \frac{1}{n}(\partial^2 {\bm \Phi}_z * p({\bf z}))^{\top}{\bf 1}.
\end{equation}
Then by taking limit over $K \to \infty$ we obtain the final RFF solution
\begin{equation}
\label{eq:rff_f}
    f_{m}^* = \lim\limits_{K \to \infty} f_K
    = \frac{1}{\lambda}{\bm \phi}(\cdot)^{\top}({\bm H} + n\lambda{\bf I})^{-1}{\bm Hh}
    - \frac{1}{\lambda}{\bm \phi}(\cdot)^{\top}{\bm h}.
\end{equation}
The detailed derivation can be found in \ref{sec:rff_solution_derivation}.

Similar result can be derived for the Nystr\"om-type approximation
(see \ref{sec:Nystrom}).
The disadvantage in this case is that we need to calculate
convolution of the first and second order derivatives with the noise distribution
for each kernel.
For RFF, on the other hand, all the terms remains the same for any shift-invariant kernel
except the distribution of weights $\bf{W}$, which is much more convenient.

Another important thing we would like to stress is that in the resulting solution
each feature has a weight proportional to
$\exp \left ( -\frac{\sigma^2}{2} \|{\bf w}_i\| \right )$
(see \ref{sec:rff_solution_derivation} for details).
This means that the high-frequency features have weight which is close to zero.
Such behaviour can be interpreted as a regularization that penalizes
oscillating terms.

There are several hyper-parameters in the approach that affects the resulting
quality, namely, the kernel hyper-parameters ${\bm \theta}$,
the regularization parameter $\lambda$ and,
assuming that the noise is zero-mean Gaussian, the noise variance $\sigma$.
To tune these parameters we use the loss on the hold-out (validation) set.
The loss in this case is ordinary score matching (no denoising) loss as we would like to estimate
how good our model approximates the original data, not the noisy one.

Another important part of the algorithm is the base density $q_0$.
From a theoretical point of view, base density is responsible for the tails of the
distribution and do not affect the estimator in the areas with high density.
Therefore, in this paper we consider three different options for $q_0$:
uniform distribution with support bounded by particular training sample,
multivariate Gaussian distribution and the mixture of Gaussians.
In the latter case, $q_0$ is fitted before training using Bayesian Mixture Model \cite{bishop}.

At the end of the training we estimate the normalizing constant via importance sampling
as was proposed in \cite{GrettonDeep}.
It should be noted that in the case of uniform base density normalization could not be
estimated properly due to the unknown data support measure.
The whole method is summarized in Algorithm \ref{alg:KDSM}.
\begin{algorithm}
\caption{Kernel denoising score matching.}
\label{alg:KDSM}
\begin{algorithmic}[1]
    \Require Training set ${\cal D}$, $m$~--- number of Fourier features,
    $n_z$~--- number of samples to estimate normalization constant,
    initial regularization parameter $\lambda$
    \State Fit $q_0({\bf x})$ to the given data set.
    \While{not stopping condition}
        \For{mini-batches ${\cal D}_t, {\cal D}_v \in {\cal D}$}
            \State Compute random Fourier approximation $f_m^*$ using equation \eqref{eq:rff_f} on ${\cal D}_t$.
            \State Compute ordinary score matching loss on validation
            \[
                \hat{J}_{val}(\lambda, \sigma, {\bm p}_k) =
                \frac{1}{|{\cal D}_v|}\left[
                    {\bf 1}^{\top}\partial^2 \bm{\Phi}_{v}\bm{b}_t +
                    \frac{1}{2}\bm{b}_t^{\top}\partial \bm{\Phi}_{v}^{\top}
                    \partial \bm{\Phi}_{v}\bm{b}_t +
                    \bm{b}_t^{\top}\partial \bm{\Phi}_{v}^{\top} \nabla\log q_0({\bf x})
                \right]
            \]
            \State Do gradient step over hyper-parameters ($\lambda$, $\sigma$, ${\bm p}_k$)
        \EndFor
    \EndWhile
    \State Compute $f_m^* = \bm{\phi}(\cdot)^{\top}\bm{b}_{\cal D}$
    using full dataset ${\cal D}$
    \State Compute normalization constant approximation $\hat{Z}$ via importance sampling
    \[
        \hat{Z} = \frac{1}{n_z}\sum_{i = 1}^{n_z}\frac{f_m^*({\bf x}_i)}{q_0({\bf x}_i)},
        \quad {\bf x}_i \sim q_0({\bf x})
    \]
\end{algorithmic}
\Return $\log p_f = f_m^* - \hat{Z}$
\end{algorithm}

The total complexity of the proposed approach is $O(m^3 + nm^2 + nmd)$,
where $O(nmd)$ operations are required generate random features,
$O(nm^2)$ is to compute feature matrix ${\bm H}$
and $O(m^3)$ corresponds to the matrix inversion which can be reduced to
$O(m^2)$ in some cases by using iterative methods for solving systems of linear equations.

Now, let us provide the bounds on the error of the approximation of the proposed approach.
Let us introduce the derivatives of the exact kernel matrix
\[
    \partial^p\partial^q {\bm K}_{(a - 1)d + i, (b - 1)d + j} = \partial^p_i\partial^q_{d + j}k({\bf x}_a, {\bf x}_b),
    \quad
    p, q \in \mathbb{N}_+.
\]
We also denote the derivatives of the random feature vector as
\[
   \partial^p\bm{\phi} = \begin{pmatrix}
        \partial_1^p\bm{\phi}({\bf w}^{\top}{\bf x}_1 + b) &
        \cdots &
        \partial^p_d\bm{\phi}({\bf w}^{\top}{\bf x}_n + b)
   \end{pmatrix}^\top.
\]
The error bounds for score matching with RFF is given by the following theorem.
\begin{theorem}
    Let $\delta \in (0, 1)$, $\varepsilon > 0$, then for
    $n \geq \frac{8}{3\varepsilon^2} \log\frac{m}{\delta}$ and assuming that
    ${\bf D}_1 = \mathbb{E}_{\bf w}{\rm tr}[\partial\bm{\phi}\partial\bm{\phi}^{\top}]
    \partial\bm{\phi}\partial\bm{\phi}^{\top} < \infty$, ${\bf D}_2 =
    \mathbb{E}_{\bf w}{\rm tr}[\partial^2\bm{\phi}\partial^2\bm{\phi}^{\top}]\partial\bm{\phi}
    \partial\bm{\phi}^{\top} < \infty$,
    we have that with probability at least $(1 - \delta)$ the following upper bound
    on distance between an averaged RFF score matching solution $f_{n, m}^*$
    and exact kernel solution $f_n^*$ holds
    \begin{align*}
        \mathbb{E}_{{\bf x}, {\bf w}}(f_{n, m}^*({\bf x}) - f_n^*({\bf x}))^2 \leq
        \frac{2}{\lambda^2 n^2 m^2}\left[
            \vphantom{\partial\partial {\bm K}^{\frac{1}{2}}} \right .
            & m\|\partial\partial {\bm K}^{\frac{1}{2}}(\partial\partial {\bm K} + \lambda n{\bf I})^{-1}\partial\partial^2 {\bm K}{\bf 1}\|^2 +\nonumber \\
            &m{\bf 1}^{\top}\partial^2\partial^2{\bm K}{\bf 1} +
            (1 + \varepsilon m)\|{\bf D}_2^{\frac{1}{2}}{\bf 1}\|^2 + \nonumber \\
            &\left . (1 + \varepsilon m)\|{\bf D}_1^{\frac{1}{2}}(\partial\partial {\bm K} + \lambda n {\bf I})^{-1}\partial\partial^2 {\bm K}{\bf 1}\|^2
        \right ].
    \end{align*}
\end{theorem}
The proof of the theorem is given in \ref{sec:error_bound_proof}.

\subsection{Discussion}

While RFF kernel approximation admits computationally efficient solution of score matching,
the convergence properties remains an open question.
Using results from \cite{Gretton2013} the convergence can be established only in the
RKHS that corresponds to the approximate kernel.
So we have the following relation
\[
    J(p_0\| p_{\lambda, n, m}) \to \inf_{p \in \Tilde{{\cal P}}}J(p_0\| p)
    \quad
    \lambda \to 0,~\lambda n \to \infty,~n \to \infty,
\]
where $p_{\lambda, n, m}$ is the density obtained using $m$ features and
$\Tilde{{\cal P}}$ is an exponential family with sufficient statistic
$\phi({\bm W}{\bf x} + {\bm b})$.
To upper bound the error of the approximation we
can consider the following inequality
\[
    \|f_{\lambda, n, m} - f_0\| \leq
        \|f_{\lambda, n, m} - f_{\lambda, m}\| +
        \|f_{\lambda, m} - f_{\lambda}\| +
        \|f_{\lambda} - f_0\|,
\]
where $f_{\lambda}$ minimizes \eqref{eq:sm_kernel} and
$f_{\lambda, n}$ is a solution of a finite sample version of \eqref{eq:sm_kernel},
$\|\cdot\|$ is a norm in $L^2({\mathbb{R}^d}, p_0)$.
$\|f_{\lambda,n, m} - f_{\lambda, m}\|$ includes the term $\|\hat{\xi}_m - \hat{\xi}\| = O(m^{-\frac{1}{2}})$ that can obtained using concentration lemma from
\cite{sutherland2017efficient} under additional assumption on the boundness
of the derivatives of the approximate kernel.
This implies that we should potentially use $O(n)$ features to obtain the same
convergence rates as in the case of exact solution.
To reduce the lower bounds on number of features
we need to provide refined analysis in a way similar to
\cite{aless2016generalization, li2018unified} in the future study.

In the case of Denoising Score Matching the estimated density will converge to the
$p^* * p_{\varepsilon}$, where $p^* = \inf_{p \in \Tilde{{\cal P}}}J(p_0\| p)$
is the density from $\Tilde{\mathcal{P}}$ closest to $p_0$.
So, on one hand the noise variance should be as small as possible.
On the other hand, the Wasserstein distance
between the approximation and the true density for the
Denoising Score Matching can be upper bounded as follows
\[
    W(p_0, p) \leq
    \mathbb{E}[\|{\bm \varepsilon}\|^2]^{\frac{1}{2}} +
    \tilde{C}\sqrt{J(p_0 * p_{\varepsilon}, p_{\lambda, n, m} * p_{\varepsilon})},
\]
where $\mathbb{E}[\|{\bm \varepsilon}\|^2] = n\sigma^2$.
The second term takes large value for small noise levels (due to different supports
of the approximate density and the true density)
and smaller values for large noise values.
So the choice of $\sigma$ is a trade-off between estimator stability and
how close it is to the unknown density function $p_0$.

\section{Results}
\label{chap:experiments}
\subsection{Experimental setup}
In all our experiments we used RBF kernel with diagonal covariance matrix,
the noise was assumed to be isotropic Gaussian (though in general
we can use arbitrary noise covariance matrix).

We compare the proposed approach (DSM RFF),
ordinary score matching with RFF (SM RFF),
exact kernel solution \eqref{eq:sm_kernel} (Exact) and
its Nystr\"om version with subsampled basis \cite{sutherland2017efficient} (Nystr\"om).
We used original implementations of this model from \cite{GrettonGit}.

The comparison is conducted on two types of data: artificially generated 2D densities
and datasets from the UCI repository \cite{UCI}
(the particular choice of data is motivated by previous research on kernel exponential family
\cite{sutherland2017efficient, Gretton2015, GrettonDeep}):
\begin{enumerate}
    \item Synthetic data generated from the following densities: a mixture of Gaussians, Uniform, Mixture on Uniforms, Cosine, Funnel, Banana, Ring, Mixture of Rings.
    \item RedWine, WhiteWine, MiniBoone.
\end{enumerate}
The Exact kernel model was not compared on MiniBoone dataset
because it's too computationally expensive.

To estimate the quality of models the following metrics were used:
\begin{enumerate}
    \item Log-likelihood (higher is better).
    It requires the normalization constant which can only be approximated, so
    the log-likelihood tends to be overestimated \cite{GrettonDeep}.
    \item Fisher divergence (lower is better).
    It requires true log-density gradient to be known and hence could be estimated only for
    artificial data, moreover, for uniform settings could be computed only on the support of
    true density.
    Alternatively, score-matching could be used, but scores for different models
    are not comparable in general.
    \item Finite-Set Stein Discrepancy (FSSD) goodness of fit test \cite{FSSD, GOF} with $0.05$
    significance level.
    We used Gaussian kernel, and its lengthscale was chosen to be median over
    pairwise distances between samples in order to avoid optimization over test points for the
    particular model, otherwise, we can not compare models.
    FSSD statistic almost surely equals to zero if and only if model density and $p_0$
    coincide.
    \item Wasserstein distance.
    In order to estimate this quantity, we used Metropolis Adjusted Langevin Algorithm (MALA)
    \cite{MALA} to draw samples from the model densities.
    We used step-size $0.1$, chain length was $10^4$ with $5\cdot 10^3$ burn-in.
\end{enumerate}

\subsection{Results}

\begin{figure}[!ht]
  \centering
  \begin{subfigure}[b]{0.32\textwidth}
    \includegraphics[width=\textwidth]{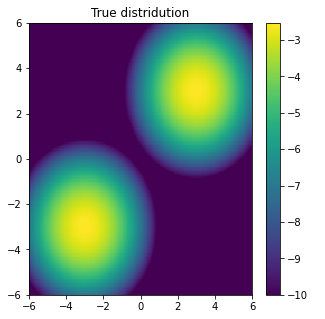}
    \captionsetup{justification=centering}
    \caption{Ground truth\\ \hfill}
    \label{sfig:MoGtrue}
  \end{subfigure}
  \begin{subfigure}[b]{0.32\textwidth}
    \includegraphics[width=\textwidth]{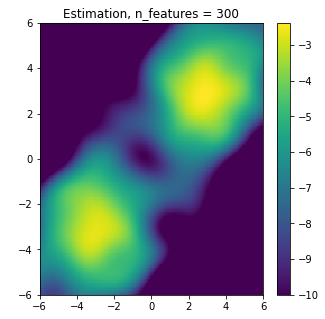}
    \captionsetup{justification=centering}
    \caption{RFF, no noise, $SM~=~-0.68$}
    \label{sfig:MoGrff}
  \end{subfigure}
  \begin{subfigure}[b]{0.32\textwidth}
    \includegraphics[width=\textwidth]{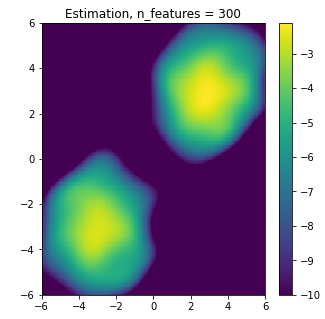}
    \captionsetup{justification=centering}
    \caption{RFF+noise, $SM~=~-0.99$}
    \label{sfig:MoGnoise}
  \end{subfigure}
  \begin{subfigure}[b]{0.32\textwidth}
    \includegraphics[width=\textwidth]{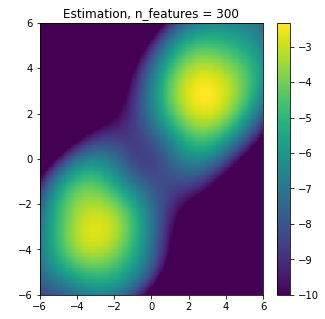}
    \captionsetup{justification=centering}
    \caption{RFF, no noise, wide kernel, $SM~=~-0.49$}
    \label{sfig:MoGrff_1}
  \end{subfigure}
  \begin{subfigure}[b]{0.32\textwidth}
    \includegraphics[width=\textwidth]{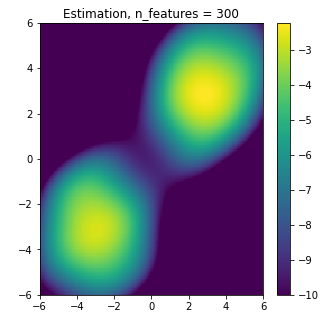}
    \captionsetup{justification=centering}
    \caption{RFF+noise, wide kernel, $SM~=~-0.67$}
    \label{sfig:MoGnoise1}
  \end{subfigure}
  \caption{Comparison of score-matching with and without noise,
    noise variance is
    $\sigma = 5\cdot10^{-4}$.
    We clip values of log-density that less then $-10$.
  }
  \label{fig:demo}
\end{figure}

We start by considering an approximate denoising approach (see \ref{sec:Taylor} for derivation)
to figure out if there is a benefit from the convolution with noise.
To accomplish this we construct illustrative experiment with $300$ RFF
features for Gaussian mixture.
We used multivariate Gaussian distribution for $q_0$
and the training set size was $10^3$.
The results are presented on Fig.~\ref{fig:demo},
from which it is clear that noisy approach better estimates ground truth in between components
region even with the presence of small noise
Also note the that there are less oscillations when we add noise to the data.

\begin{figure}[!ht]
    \centering
    \begin{subfigure}[b]{\textwidth}
      \includegraphics[width=\textwidth]{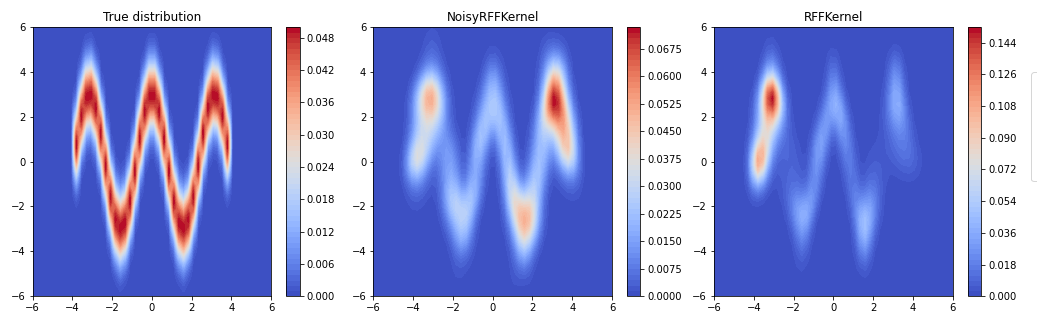}
      \caption{Cosine}
      \label{sfig:1000Cos}
    \end{subfigure}
    \begin{subfigure}[b]{\textwidth}
      \includegraphics[width=\textwidth]{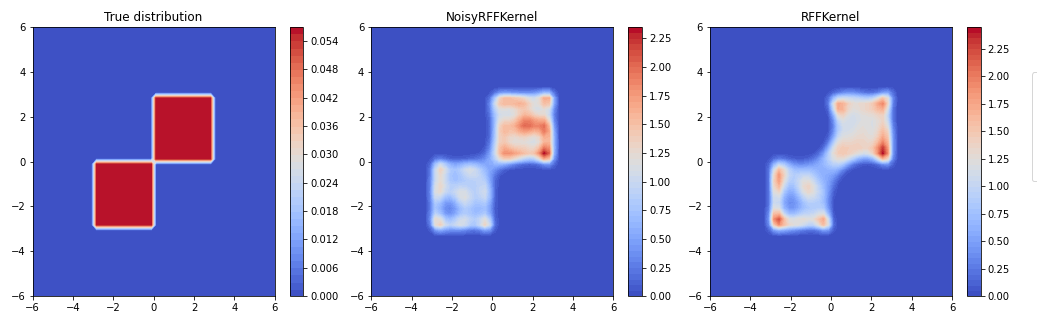}
      \caption{Mixture of uniforms}
      \label{sfig:1000Rings}
    \end{subfigure}
    \caption{Density estimates using DSM RFF (middle column) and SM RFF (right column).
    The ground truth density is in the first column.
    }
    \label{fig:2d_1000}
\end{figure}

The next step is to compare the proposed algorithm to other approaches on synthetic 2D data
and datassets from UCI.
In this case we used $512$ Random Fourier Features.
As the base density $q_0$ we used mixture of Gaussians.
As in the previous example a relatively small sample size was used.
Our models were trained for $60$ iterations using Adam optimizer with $0.1$ learning rate,
$512$ features were used.
The results for Cosine and mixture of uniforms are presented
in Figure~\ref{fig:2d_1000} (other results can be find in \ref{sec:C}).
For cosine data the form of distribution estimated via DSM RFF is much closer
to the real one.
However, for the mixture of uniforms it fails to correctly estimate weights
of the components.
In both of this densities, we observe model misspecification in the case of SM RFF
because all other 2D densities have full space support.
For the rest distributions there is no significant visual difference.

These experiments showed that denoising score matching with RFF
in general works better for distributions with bounded support.
For the multimodal distributions it could fail to correctly estimate weights of
components or oversmooth the areas between components.
\begin{figure}[!ht]
  \centering
  \begin{subfigure}[b]{0.32\textwidth}
    \includegraphics[width=\textwidth]{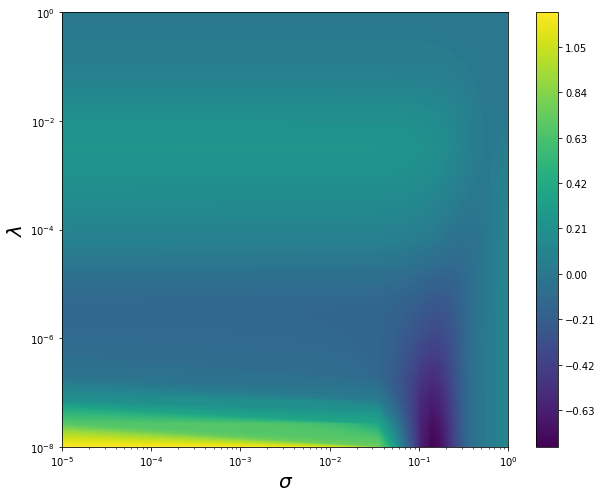}
    \caption{Uniform}
  \end{subfigure}
  \begin{subfigure}[b]{0.32\textwidth}
    \includegraphics[width=\textwidth]{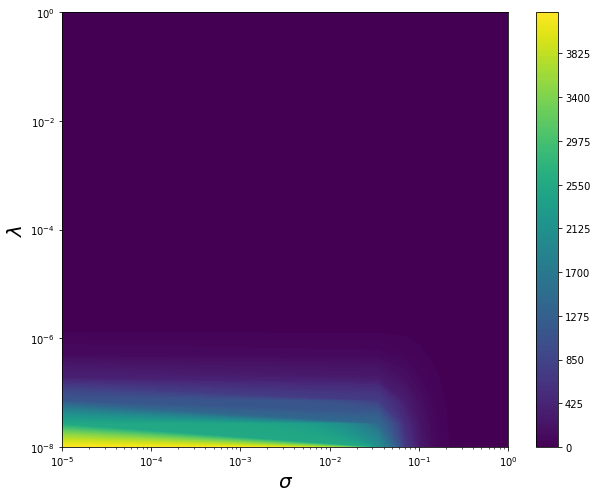}
    \caption{Funnel}
  \end{subfigure}
  \begin{subfigure}[b]{0.32\textwidth}
    \includegraphics[width=\textwidth]{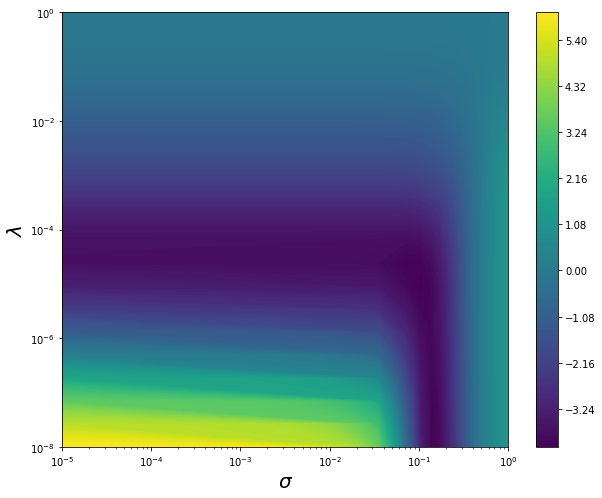}
    \caption{Two Rings}
  \end{subfigure}
  \caption{Dependence of loss on the regularization parameter $\lambda$ (y axis) and noise $\sigma$ (x axis).}
  \label{fig:lossdemo}
\end{figure}
Another observation about the approach is that in some cases it tends to choose
large noise variance.
In Fig.~\ref{fig:lossdemo} we visualize the dependence of the loss on
the regularization parameter and noise variance for several 2D distributions.
Interestingly, for "good" distributions
(like Funnel, that have full space support and one mode) the loss surface has wide
minimum w.r.t regularization and noise variance.
For multimodal distributions the loss surface has narrower minimum.
For uniform distribution (which differs from other that it has bounded support)
the minimum w.r.t noise variance is narrow but it is also separated from zero.
This indicates the need for the noise in such cases.

\begin{figure}[h]
  \centering
  \includegraphics[width=\textwidth]{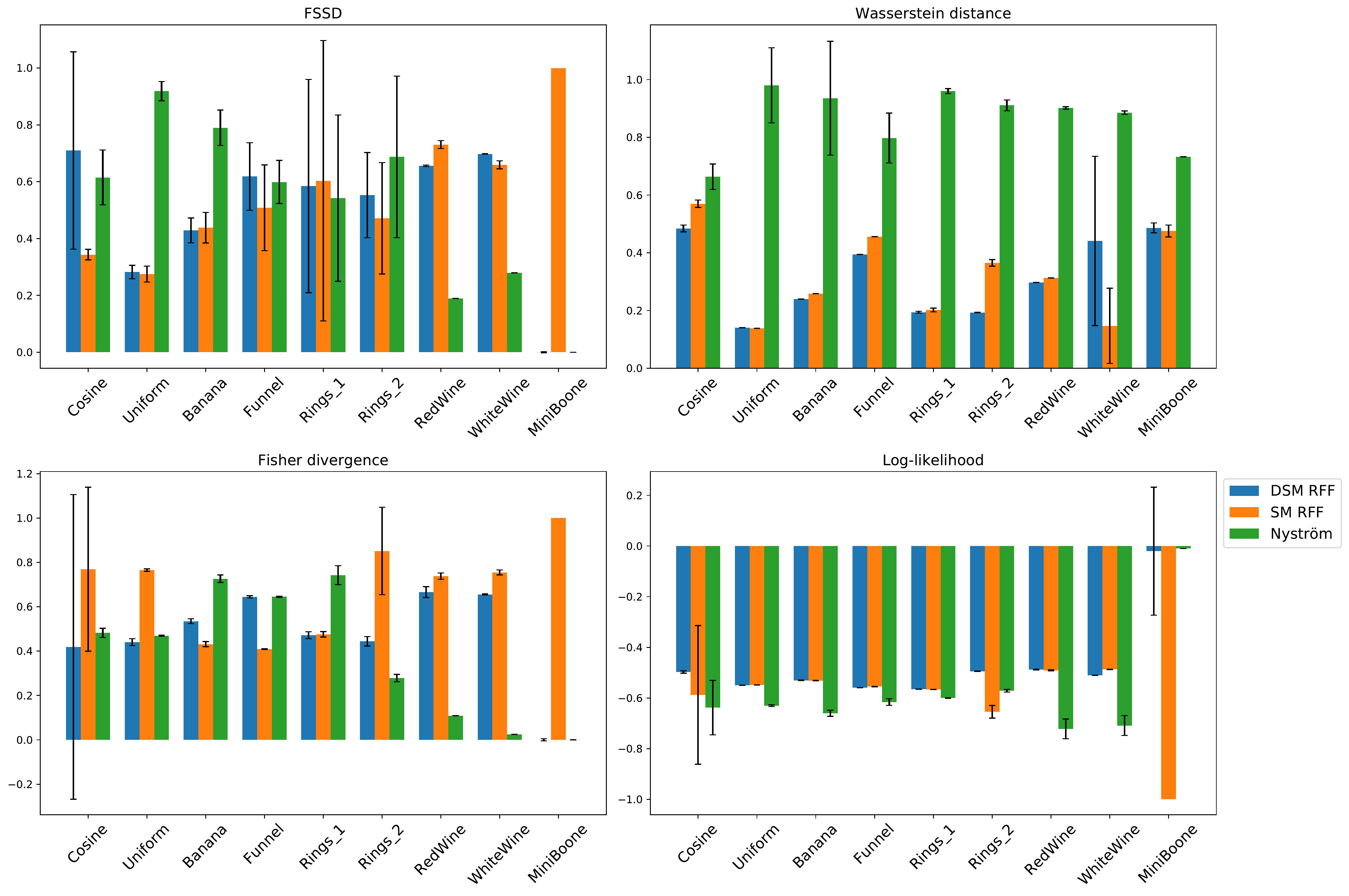}
  \caption{Metrics on different datasets for different methods.
           For each dataset each metric was normalized across methods
           to have unit norm.
           We did it only for better visualization.}
  \label{fig:metrics}
\end{figure}

In Figure~\ref{fig:metrics} we plot all the metrics for all data sets.
For each dataset each metric was normalized across methods to have unit norm.
This was done only for better visualization.
The original values are given in \ref{sec:B}.
The figure illustrates the mean value of the metrics and corresponding variance
calculated across $10$ runs.
From the figure we can see, that w.r.t. almost all metrics (except the log-likelihood)
the proposed approach shows better or comparable results in many cases.
Actually, the Wasserstein distance is smaller for DSM RFF for all data sets.
We can also see, that SM RFF tends to have larger variance than its noisy version.

In Table~\ref{tab:metrics} we provide results for the datasets from the UCI repository
as well as the training time.
The MiniBoone dataset is large and the Nystr\"om-based implementation
could not fit into memory, so we had to train the model using only a subset of
$15000$ samples.
Other methods were trained using the whole data set.
To fairly compare the training time the experiments were conducted on
Intel(R) Core(TM) i7-7820X CPU @ 3.60GHz with 64Gb RAM.
We can see that the proposed approach is much faster than the implementation
of the Nystr\"om based approach.

\begin{table}
\centering
\caption{Metrics for the data sets from UCI repository.}
\label{tab:metrics}
  \footnotesize
  \begin{tabular}{llrrrrr}
    \toprule
    Data set & Model & Log-likelihood & FSSD &
    \shortstack{Wasserstein \\ distance} & time, s \\
    \midrule
    RedWine & DSM RFF     & -11.64 & 0.38 & 0.24 & 62 \\
            & SM RFF      & -11.72 & 0.43 & 0.25 & 61 \\
            & Nystr\"om   & -17.23 & 0.11 & 0.73 & $0.2 \times 10^4$ \\
    WhiteWine & DSM RFF   & -12.81 & 0.57 & 0.33 & 180 \\
              & SM RFF    & -12.22 & 0.53 & 0.11 & 180 \\
              & Nystr\"om & -17.79 & 0.23 & 0.67 & $1 \times 10^4$ \\
    MiniBoone & DSM RFF   & -93.11 & 307.67 & 0.49 & $0.5 \times 10^4$ \\
              & SM RFF    & -4580.20 & $2 \times 10^8$ & 0.48 & $0.5 \times 10^4$ \\
              & Nystr\"om & -46.06 & 0.02 &  0.75 & $0.6 \times 10^4$ \\
    \bottomrule
  \end{tabular}
\end{table}

\section{Conclusion}
\label{chap:Conclusion}
In this work we presented denoising score matching for the kernel exponential family.
The computational complexity issue was approached using Random Fourier Features technique.
We derived a closed-form solution that more accurate estimate of the loss in denoising score
matching.
The proposed approach is also computationally more efficient than existing
approaches to model kernel exponential family using Nystr\"om-type approximation.
The obtained solution naturally regularizes the complexity of the model
due to the convolution with the noise which can also be interpreted as
an additional regularization parameter.
The analytical expression also allows to tune the the noise parameters
as it is now explicitly present in the expression.
So, we can use gradient based methods to optimize over noise parameters.
Convolution with noise prevents model misspecification and allows to build accurate models
in case true density lies in a lower-dimensional manifold.

The obtained model was tested on synthetic and real world datasets.
Our experiments showed that additional noise is reasonable for complex multimodal distributions
or distributions with bounded support.
The proposed model give better estimate in-between modes.
Empirical study of the loss surface revealed the need to use the noise,
especially for the misspecified case.
We also provide the bounds for the method, however, they are not tight
and refining the results is planned for the future work.

\bibliographystyle{elsarticle-num}
\bibliography{main}
\appendix
\section{Technical Results}
\label{chap:Technical}

\subsection{Exact solution for the Kernel Denoising Score Matching with RFF}
\label{sec:exact_solution}
We start with the first order optimality condition:
\begin{equation*}
    \int p_{\varepsilon}({\bf y}) A^*({\bf y})\nabla V(A({\bf y})f)d{\bf y} + \lambda f = 0,
\end{equation*}
where $A^*({\bf y}): \mathbb{R}^m \to \mathcal{H}$ is an adjoint to $A({\bf y})$ and
$A^*({\bf y}){\bm \alpha} = \sum_{i = 1}^m \alpha_i\phi_i({\bf y}, \cdot)$.
Denoting
\begin{align*}
    &{\bm \alpha}({\bf y}) = -\frac{1}{\lambda}\nabla V(A({\bf y})f),\quad f = \int p_{\varepsilon}({\bf y})A^*({\bf y}){\bm \alpha}({\bf y})d {\bf y},
\end{align*}
the first order optimality condition could be rewritten as an integral equation on
${\bm \alpha}({\bf y})$:
\begin{equation}
    {\bm \alpha}({\bf y}) = -\frac{1}{\lambda}\nabla V\left(\int p_{\varepsilon}({\bf z})A({\bf y})A^*({\bf z})\alpha({\bf z})d{\bf z}\right),
\end{equation}
where $A({\bf y})A^*({\bf z})\alpha({\bf z}) = \sum\limits_{i = 1}^m\alpha_i({\bf z})\{\langle \phi_j({\bf y}, \cdot),  \phi_i({\bf z}, \cdot)\rangle\}_{j = 1}^m = {\bm K}({\bf y}, {\bf z}){\bm \alpha}({\bf z})$.

The gradient of $V$ is given
by~$\nabla V = (\frac{1}{n}, \frac{1}{n}, \ldots, \frac{1}{n}, 1)^{\top}$.
Then, we have ${\bm \alpha}({\bf y}) = ({\bm \beta}^{\top}({\bf y}), \delta)^{\top}$,
where $\delta = -\frac{1}{\lambda}$.
The integral equation on ${\bm \beta}({\bf y})$ can be expressed as
\begin{equation}
    {\bm \beta}({\bf y}) = -\frac{1}{n\lambda}\int p_{\varepsilon}({\bf z})\hat{A}({\bf y})\hat{A}({\bf z})^*{\bm \beta}({\bf z})d{\bf z} + \frac{1}{n\lambda^2}\int p_{\varepsilon}({\bf z}) \hat{A}({\bf y}) \phi_m({\bf z}, \cdot)d{\bf z},
\end{equation}
where $(\hat{A}({\bf y})f)_i = (A({\bf y})f)_i$, $i = 1, \ldots, m - 1$.
Let $b = \int p_{\varepsilon}({\bf y})\phi_m({\bf y}, \cdot) d {\bf y}$ and
$C = \int p_{\varepsilon}({\bf y})\hat{A}^*({\bf y}){\bm \beta}({\bf y}) d {\bf y}$.
Then we search for the solution of \eqref{eq:beta_integal} in the form
${\bm \beta}({\bf y}) = -\frac{1}{n\lambda}\hat{A}({\bf y})C + \frac{1}{n\lambda^2}\hat{A}({\bf y})b$.
In this case we have
\begin{equation*}
    \hat{A}({\bf y})\left[C + \frac{1}{n\lambda}BC - \frac{1}{n\lambda^2}Bb\right] = 0,
\end{equation*}
where $B = \int p_{\varepsilon}({\bf y})\hat{A}^*({\bf y})\hat{A}({\bf y}) d{\bf y}$ and
$b \in \mathcal{H}$ is a convolution of $\phi_m$ and noise density $p_{\varepsilon}$.
% Indeed, fixing ${\bf z}$, $\phi_m({\bf y, z}) \in H$ as a function of ${\bf y}$. Then ${\cal F}[\phi_m({\bf y, z}) * p_{\varepsilon}({\bf y})] = {\cal F}[\phi_m({\bf y, z})]{\cal F}[p_{\varepsilon}({\bf y})]$.
% $$\int \frac{|{\cal F}[\phi_m({\bf y, z})]{\cal F}[p_{\varepsilon}({\bf y})]|^2}{{\cal F}[k](w)}dw \leq \sup|{\cal F}[p_{\varepsilon}({\bf y})]|^2\int \frac{|{\cal F}[\phi_m({\bf y, z})]|^2}{{\cal F}[k](w)}dw < \infty$$
% The last follow $p_{\varepsilon}({\bf y}) \in L^1$.

%As produnct of function from $H$ belongs to $H$
Solution $C^*$ of the above equation provides ${\bm \beta}^*({\bf y})$ and, as a result,
the solution to the initial problem.
Let us show that the obtained estimator belongs to $H$.
In fact, since
\[
    C + \frac{1}{n\lambda}BC - \frac{1}{n\lambda^2}Bb \in {\rm Ker}\hat{A}({\bf y}) \subseteq
    \mathcal{H}
\]
and $B + n\lambda I$ is continuously invertible we have that $C^* \in \mathcal{H}$.
Finally, we have
\[
    f^* = B\left[-\frac{1}{n\lambda}C^* + \frac{1}{n\lambda^2}b\right] = C^* - \frac{1}{\lambda}b - \gamma \in \mathcal{H},
\]
where we assume $\gamma \in {\rm Ker}\hat{A}({\bf y})$.

\subsection{RFF solution derivation}
\label{sec:rff_solution_derivation}
Let us use the expressions for the solution without noise
(here for simplicity the term with $\partial_i \log q_0({\bf x}_a)$ is omitted):
\begin{equation*}
    \left [ \hat{A}({\bf 0})\hat{A}({\bf 0})^* \right ]_{(a - 1)d + i, (b - 1)d + j} =
    \partial_i\partial_{j + d}k({\bf x}_a, {\bf x}_b),~\quad a, b \in [n],~i, j \in [d],
\end{equation*}
\begin{equation*}
    \left [ \hat{A}({\bf 0}){\bm \phi}_m({\bf 0}, \cdot) \right ]_{(a - 1)d + i} =
    \frac{1}{n}\sum_{b, j = 1}^{n, d}\partial_i\partial^2_{j + d}k({\bf x}_a, {\bf x}_b),~\quad a\in [n],~i \in [d].
\end{equation*}
Then we have
$\hat{A}({\bf y})\hat{A}({\bf z})^* \approx \partial{\bm \Phi}_y\partial{\bm \Phi}_z^{\top}$
and
$\hat{A}({\bf y}){\bm \phi}_m({\bf z}, \cdot) * p_{\varepsilon}({\bf z}) \approx \frac{1}{n}
\partial{\bm \Phi}_y(\partial^2{\bm \Phi}_z * p_{\varepsilon}({\bf z}))^{\top}{\bf 1}$.

Now we have everything to obtain RFF approximation of \eqref{eq:beta_finite_sample}:
\begin{align*}
    {\bm \beta}_K = -\frac{1}{nK\lambda}\partial {\bm \Phi}_K\partial {\bm \Phi}_K^{\top}{\bm \beta}_K  + \frac{1}{n^2\lambda^2}\partial{\bm \Phi}_K \odot (\partial^2 {\bm \Phi}_z * p({\bf z}))^{\top}{\bf 1},
\end{align*}
where
\[  \partial {\bm \Phi}_K =
    \begin{bmatrix}
        {\bm \Phi}_{z_1}\\
        \vdots\\
        {\bm \Phi}_{z_K}
    \end{bmatrix}, \quad
    \partial{\bm \Phi}_K \odot (\partial^2 {\bm \Phi}_z * p({\bf z}))^{\top}{\bf 1} =
    \begin{bmatrix}
        \partial{\bm \Phi}_{z_1}(\partial^2 {\bm \Phi}_z * p({\bf z}))^{\top}{\bf 1}\\
        \vdots\\
        \partial{\bm \Phi}_{z_K}(\partial^2 {\bm \Phi}_z * p({\bf z}))^{\top}{\bf 1}
    \end{bmatrix}.
\]
Denoting
\begin{equation*}
    {\bm h} = \frac{1}{n}(\partial^2 {\bm \Phi}_z * p({\bf z}))^{\top}{\bf 1},
    \quad
    {\bm H} =
    \int p_{\varepsilon}({\bf y})\partial{\bm \Phi}_y^{\top}\partial{\bm \Phi}_y d{\bf y}
\end{equation*}
we obtain the following expression for the descretized solution $f_K$:
\begin{align*}
    f_K
    &= \frac{1}{n\lambda^2}{\bm \phi}(\cdot)^{\top}{\bm H} \left[-\frac{1}{K}\partial {\bm \Phi}_K^{\top}\left(\frac{1}{K}\partial {\bm \Phi}_K\partial {\bm \Phi}_K^{\top} + n\lambda {\bf I}\right)^{-1}\partial{\bm \Phi}_K \odot {\bm h} + {\bm h}\right]
   -\frac{1}{\lambda} {\bm \phi}(\cdot)^{\top}{\bm h}\\
    &= \frac{1}{n\lambda^2}{\bm \phi}(\cdot)^{\top}{\bm H}
    \left[-\frac{1}{K}\left(\frac{1}{K}\partial  {\bm \Phi}_K^{\top}\partial{\bm \Phi}_K + n\lambda {\bf I}\right)^{-1}\partial {\bm \Phi}_K^{\top}\partial{\bm \Phi}_K \odot {\bm h} + {\bm h}\right]
    -\frac{1}{\lambda} {\bm \phi}(\cdot)^{\top}{\bm h}.
\end{align*}
By taking a limit over $K \to \infty$, and using
${\bm H} = \lim\limits_{K \to \infty}\frac{1}{K}\partial {\bm \Phi}_K^{\top}\partial{\bm\Phi}_K$ alond with
\begin{equation*}
    \lim\limits_{K \to \infty}(n\lambda{\bf I} + \frac{1}{K}\partial {\bm \Phi}^{\top}_K\partial {\bm \Phi}_K) \lim\limits_{K \to \infty}\frac{1}{K}\partial {\bm \Phi}^{\top}_K{\bm \beta}_K = \lim\limits_{K \to \infty}\frac{1}{K\lambda}\partial {\bm \Phi}^{\top}_K\partial {\bm \Phi}_K{\bm h}
\end{equation*}
the solution $f^*$ is given:
\begin{align*}
    f_{m}^* = \lim\limits_{K \to \infty} f_K
    &= -\frac{1}{n\lambda^2}{\bm\phi}(\cdot)^{\top}{\bm H}({\bm H} +
    n \lambda {\bf I})^{-1}{\bm Hh} + \frac{1}{n\lambda^2}{\bm \phi}(\cdot)^{\top}{\bm Hh} -
    \frac{1}{\lambda}{\bm \phi}(\cdot)^{\top}{\bm h}\nonumber\\
    &= \frac{1}{\lambda}{\bm \phi}(\cdot)^{\top}({\bm H} + n\lambda{\bf I})^{-1}{\bm Hh}
    - \frac{1}{\lambda}{\bm \phi}(\cdot)^{\top}{\bm h},
\end{align*}
where index $m$ refers to the number of RFF features.

\subsection{Proof for the error bounds of score matching with RFF}
\label{sec:error_bound_proof}
The idea of the proof is to upper bound the expected square difference between solutions:
\begin{equation}
    \mathbb{E}_{{\bf x}, {\bf w}}(f_{n, m}^*({\bf x}) - f_n^*({\bf x}))^2,
\end{equation}
where the difference between RFF and exact kernel solutions ($f_{n, m}^*,~f_n^*$)
is expressed as follows:
\begin{align*}
    f_{n, m}^* - f_n^*
    &= -\frac{1}{\lambda n}\partial^2{\bm k}(\cdot)^{\top}{\bf 1}
    + \frac{1}{\lambda n} {\bm \phi}^{\top}(\cdot)\partial^2\bm{\Phi}^{\top}{\bf 1} \\
    &+ \frac{1}{\lambda n}\partial{\bm k}(\cdot)^{\top}(\partial\partial {\bm K} +
    \lambda n{\bf I})^{-1}\partial\partial^2 {\bm K}{\bf 1}
    -\frac{1}{\lambda n}{\bm \phi}^{\top}(\cdot)\partial\bm{\Phi}^{\top}(\partial\bm{\Phi}
    \partial\bm{\Phi}^{\top} +
    \lambda n {\bf I})^{-1}\partial\bm{\Phi}\partial^2\bm{\Phi}^{\top}{\bf 1} \\
    &= \frac{1}{\lambda n}(\partial^2\bm{\Phi}{\bm \phi}(\cdot) -
    \partial^2{\bm k}(\cdot))^{\top}{\bf 1}
    + \frac{1}{\lambda n}(\partial{\bm k}(\cdot) -
    \partial\bm{\Phi}{\bm \phi}(\cdot))^{\top}(\partial\partial {\bm K} +
    \lambda n{\bf I})^{-1}\partial\partial^2 {\bm K}{\bf 1} \\
    &+ \frac{1}{\lambda n}{\bm \phi}^{\top}(\cdot)\partial\bm{\Phi}^{\top} \left[
        (\partial\partial {\bm K} + \lambda n{\bf I})^{-1} -
        (\partial\bm{\Phi}\partial\bm{\Phi}^{\top} + \lambda n {\bf I})^{-1}
    \right] \partial\partial^2 {\bm K}{\bf 1} \\
    &+ \frac{1}{\lambda n}{\bm \phi}^{\top}(\cdot)\partial\bm{\Phi}^{\top}(\partial\bm{\Phi}
    \partial\bm{\Phi}^{\top} + \lambda n {\bf I})^{-1}(\partial\partial^2 {\bm K} -
    \partial\bm{\Phi}\partial^2\bm{\Phi}^{\top}){\bf 1}.
\end{align*}
The above expectation is taken jointly over random Fourier weights and given points
${\bf x} \sim p_0$.
It can be written as
$\mathbb{E}_{{\bf x}, {\bf w}}[f] = \mathbb{E}_{{\bf w}}\mathbb{E}_{{\bf x}}[f | {\bf w}]$.
The first term in the above expression is the difference between $\hat{\xi}$
and its RFF approximation $\hat{\xi}_m$, so, we have:
\begin{align*}
    \mathbb{E}_{{\bf w}}{\bf 1}^{\top}&(\partial^2\bm{\Phi}{\bm \phi}(\cdot) - \partial^2{\bm k}(\cdot))(\partial^2\bm{\Phi}{\bm \phi}(\cdot) - \partial^2{\bm k}(\cdot))^{\top}{\bf 1} \\
    &= {\bf 1}^{\top}\left[\mathbb{E}_{{\bf w}}[\partial^2\bm{\Phi}{\bm \phi}(\cdot){\bm \phi}(\cdot)^{\top}\partial^2\bm{\Phi}^{\top}] - \partial^2{\bm k}(\cdot)\partial^2{\bm k}(\cdot)^{\top}\right]{\bf 1}\\
    &\leq \frac{m - 1}{m}{\bf 1}^{\top}\partial^2{\bm k}(\cdot)\partial^2{\bm k}^{\top}(\cdot){\bf 1} + \frac{1}{m}{\bf 1}^{\top}\partial^2\partial^2{\bm K}{\bf 1} - {\bf 1}^{\top}\partial^2{\bm k}(\cdot)\partial^2{\bm k}(\cdot)^{\top}{\bf 1}\\
    &\leq \frac{1}{m}{\bf 1}^{\top}\partial^2\partial^2{\bm K}{\bf 1},
\end{align*}
where the first inequality is obtained using
$\sup_{\bf x}|\phi_i({\bm W}{\bf x} + {\bm b})| \leq 1$.
As this expression does not depend on ${\bf x}$, the joint expectation will be the same.

For the second term in $f_{n, m}^*({\bf x}) - f_n^*({\bf x})$ derivation of the
upper bound is technically the same, but with lower-order derivatives, so
\begin{align*}
    \mathbb{E}_{{\bf w}}\left[(\partial{\bm k}(\cdot) -
    \partial\bm{\Phi}{\bm \phi}(\cdot))^{\top}(\partial\partial {\bm K} +
    \lambda n{\bf I})^{-1}\partial\partial^2 {\bm K}{\bf 1}\right]^2 \leq \\
    \frac{1}{m}\|\partial\partial {\bm K}^{\frac{1}{2}}(\partial\partial {\bm K} + \lambda n{\bf I})^{-1}\partial\partial^2 {\bm K}{\bf 1}\|^2.
\end{align*}

Third them:
\begin{align*}
    \mathbb{E}_{{\bf x, w}}&\left[{\bm \phi}^{\top}(\cdot)\partial\bm{\Phi}^{\top}\left[(\partial\partial {\bm K} + \lambda n{\bf I})^{-1} - (\partial\bm{\Phi}\partial\bm{\Phi}^{\top} + \lambda n {\bf I})^{-1}\right]\partial\partial^2 {\bm K}{\bf 1}\right]^2 \\
    &= \mathbb{E}_{{\bf x, w}}\left[{\bm \phi}^{\top}(\cdot)\partial\bm{\Phi}^{\top}(\partial\bm{\Phi}\partial\bm{\Phi}^{\top} + \lambda n{\bf I})^{-1}(\partial\partial {\bm K} - \partial\bm{\Phi}\partial\bm{\Phi}^{\top})(\partial\partial {\bm K} + \lambda n {\bf I})^{-1}\partial\partial^2 {\bm K}{\bf 1}\right]^2\\
    &\leq \mathbb{E}_{{\bf w}}\mathbb{E}_{{\bf x}}\left[\|{\bf R}\|_2 \|(\partial\partial {\bm K} - \partial\bm{\Phi}\partial\bm{\Phi}^{\top})(\partial\partial {\bm K} + \lambda n {\bf I})^{-1}\partial\partial^2 {\bm K}{\bf 1}\|^2 | {\bf w}\right],
\end{align*}
where only $\bf R$ depends on ${\bf x}$.
\begin{align*}
    \mathbb{E}_{\bf x}{\bf R}
    &= (\partial\bm{\Phi}\partial\bm{\Phi}^{\top} + \lambda n{\bf I})^{-1}\partial\bm{\Phi}
    \mathbb{E}_{\bf x}\left[{\bm \phi}(\cdot){\bm \phi}^{\top}(\cdot)\right]\partial\bm{\Phi}^{\top}(\partial\bm{\Phi}\partial\bm{\Phi}^{\top} + \lambda n{\bf I})^{-1}\\
    &= \frac{1}{n} (\partial\bm{\Phi}\partial\bm{\Phi}^{\top} + \lambda n{\bf I})^{-1}\partial\bm{\Phi}
    (\bm{\Phi}^{\top}\bm{\Phi} + \varepsilon n I)\partial\bm{\Phi}^{\top}(\partial\bm{\Phi}\partial\bm{\Phi}^{\top} + \lambda n{\bf I})^{-1}
\end{align*}
This inequality holds with probability $1 - \delta$ for
$n \geq \frac{8}{3\varepsilon^2} \log\frac{m}{\delta}$
%$n \geq \frac{8 \sup_{\bf x}k({\bf x}, {\bf x})}{3\varepsilon^2} \log\frac{m}{\delta}$
and obtained from the Bernstein inequality assuming that the weights are fixed \cite{Tropp_2015}.
\begin{align*}
    \lambda_{\max}({\bf R})
    &= \frac{1}{n}\lambda_{\max}\left[(\partial\bm{\Phi}\partial\bm{\Phi}^{\top} +
    \lambda n{\bf I})^{-1}\partial\bm{\Phi}
    (\bm{\Phi}^{\top}\bm{\Phi} + n\varepsilon I)\partial\bm{\Phi}^{\top}(\partial\bm{\Phi}
    \partial\bm{\Phi}^{\top} + \lambda n{\bf I})^{-1}\right]\\
    &\leq \frac{1}{n}\lambda_{\max}\left[(\bm{\Phi}^{\top}\bm{\Phi} + n\varepsilon I)
    \partial\bm{\Phi}^{\top}(\partial\bm{\Phi}\partial\bm{\Phi}^{\top} + \lambda n{\bf I})^{-1}\partial\bm{\Phi}
    \right]\\
    &\leq \frac{1}{n}\lambda_{\max}\left[(\bm{\Phi}^{\top}\bm{\Phi} + n\varepsilon I)\right] \leq
    \frac{1}{n}{\rm tr}\left[(\bm{\Phi}^{\top}\bm{\Phi} + n\varepsilon I)\right] \\
    &\leq \left(\frac{1}{m}\max_i\sup_{\bf x}\|\phi_i({\bm W}{\bf x} + {\bm b})\|^2 +
    \varepsilon\right) \leq \frac{1}{m} + \varepsilon
\end{align*}
\begin{align*}
    \mathbb{E}_{\bf w}&\|(\partial\partial {\bm K} - \partial\bm{\Phi}\partial\bm{\Phi}^{\top})(\partial\partial {\bm K} + \lambda n {\bf I})^{-1}\partial\partial^2 {\bm K}{\bf 1}\|^2\\
    & = {\bf 1}^{\top}\partial\partial^2 {\bm K}^{\top}(\partial\partial {\bm K} + \lambda n {\bf I})^{-1}(\mathbb{E}_{\bf w}\partial\bm{\Phi}\partial\bm{\Phi}^{\top}\partial\bm{\Phi}\partial\bm{\Phi}^{\top} - \partial\partial {\bm K}^2)(\partial\partial {\bm K} + \lambda n {\bf I})^{-1}\partial\partial^2 {\bm K}{\bf 1}
\end{align*}
\begin{align*}
    \mathbb{E}_{\bf w}\partial\bm{\Phi}\partial\bm{\Phi}^{\top}\partial\bm{\Phi}\partial\bm{\Phi}^{\top} = \frac{m - 1}{m} \partial\partial {\bm K}^2 + \frac{1}{m}{\bf D}_1
\end{align*}
where the latter term is obtained under assumption that ${\bf D}_1$ does not depend on
${\bf x}$.
This assumption holds for a sufficiently smooth kernels and we can
rewrite the expression under an expectation as polynomial of weights times
trigonometric function.
\begin{align*}
    \mathbb{E}_{\bf w}&\|(\partial\partial {\bm K} - \partial\bm{\Phi}\partial\bm{\Phi}^{\top})(\partial\partial {\bm K} + \lambda n {\bf I})^{-1}\partial\partial^2 {\bm K}{\bf 1}\|^2
    \leq \frac{1}{m}\|{\bf D}_1^{\frac{1}{2}}(\partial\partial {\bm K} + \lambda n {\bf I})^{-1}\partial\partial^2 {\bm K}{\bf 1}\|^2
\end{align*}
Analogously, for the last term under assumption that ${\bf D}_2 < \infty$ we have
\begin{align*}
    \mathbb{E}_{{\bf x, w}} \partial\bm{\Phi}\partial^2\bm{\Phi}^{\top}\partial\bm{\Phi}\partial^2\bm{\Phi}^{\top} \leq \frac{1}{m}\|{\bf D}_2^{\frac{1}{2}}{\bf 1}\|^2.
\end{align*}

Finally, combining all the above, we have
\begin{align*}
    \mathbb{E}_{{\bf x}, {\bf w}}(f_{n, m}^*({\bf x}) - f_n^*({\bf x}))^2 \leq \frac{2}{\lambda^2 n^2 m^2}\left[
    m{\bf 1}^{\top}\partial^2\partial^2{\bm K}{\bf 1} + m\|\partial\partial {\bm K}^{\frac{1}{2}}(\partial\partial {\bm K} + \lambda n{\bf I})^{-1}\partial\partial^2 {\bm K}{\bf 1}\|^2 \right. \nonumber \\
    \left. + (1 + \varepsilon m)\|{\bf D}_2^{\frac{1}{2}}{\bf 1}\|^2 + (1 + \varepsilon m)\|{\bf D}_1^{\frac{1}{2}}(\partial\partial {\bm K} + \lambda n {\bf I})^{-1}\partial\partial^2 {\bm K}{\bf 1}\|^2
    \right].
\end{align*}

\subsection{Derivation of ${\bm H}$ and  ${\bm h}$ \eqref{eq:h} for Gaussian noise}
\label{sec:Hh_der}
    \begin{align*}
        {\bm H}
        &= \partial \Phi^{\top}\partial \Phi * p_{\varepsilon}\\
        &= \sum\limits_{a = 1}^n\sum\limits_{i = 1}^d \partial_i \phi(\bm{W}{\bf x}_a + \bm{b})\partial_i \phi^{\top}(\bm{W}{\bf x}_a + \bm{b}) * p_{\varepsilon}\\
        &= \sum\limits_{a = 1}^n\sum\limits_{i = 1}^d \bm{W}_{:, i}\bm{W}_{:, i}^\top \odot \phi'(\bm{W}{\bf x}_a + \bm{b}) \phi'^\top(\bm{W}{\bf x}_a + \bm{b}) * p_{\varepsilon}\\
        &= \frac{1}{M}\bm{W}\bm{W}^{\top} \odot \sum\limits_{a = 1}^n \sin(\bm{W}{\bf x}_a + \bm{b}) \sin^\top(\bm{W}{\bf x}_a + \bm{b}) * p_{\varepsilon}
        \label{eq:H_derivation}
    \end{align*}
    Assuming that $p_{\varepsilon} = {\cal N}({\bf 0}, \sigma^2{\bf I})$ and using
    \begin{align*}
        \cos(\mathbf{w}^\top \mathbf{x}) * \mathcal{N}(0, \sigma^2\mathbf{I}) &=
        %\int_\mathbf{y} \cos(\mathbf{w}^\top(\mathbf{x} - \mathbf{y}))
        %\frac{\exp\left(-\frac{\|\mathbf{y}\|^2}{2\sigma^2}\right )}{Z} d\mathbf{y}
         e^{-\frac{\sigma^2}{2}\|\mathbf{w}\|_2^2}\cos(\mathbf{w}^\top \mathbf{x})
    \end{align*}
    we will obtain:
    \begin{align*}
        \sin(\mathbf{w^\top x} + b) \sin(\mathbf{v^\top x} + c) * p_{\varepsilon}
        &= \frac{1}{2}\left[\cos ((\mathbf{w - v})^\top \mathbf{x} + b - c) - \cos ((\mathbf{w + v} + b + c)^\top \mathbf{x})\right] * p_{\varepsilon}\\
        &= \frac{1}{2}e^{-\frac{\sigma^2}{2}\|\mathbf{w - v}\|_2^2}\cos ((\mathbf{w - v})^\top \mathbf{x} + b - c)\\
        &- \frac{1}{2}e^{-\frac{\sigma^2}{2}\|\mathbf{w + v}\|_2^2}\cos ((\mathbf{w + v})^\top \mathbf{x} + b + c)
    \end{align*}
    \begin{align}
        {\bm H} = \frac{1}{2M}\bm{W}\bm{W}^{\top} \odot \sum\limits_{a = 1}^n\left[
            e^{-\frac{\sigma^2}{2}\|{\bf w}_i - {\bf w}_j\|_2^2}\cos (({\bf w}_i - {\bf w}_j)^\top \mathbf{x}_a + {\bf b}_i - {\bf b}_j)
            \right .\nonumber \\
            \left .
            - e^{-\frac{\sigma^2}{2}\|{\bf w}_i + {\bf w}_j\|_2^2}\cos (({\bf w}_i + {\bf w}_j)^\top \mathbf{x}_a + {\bf b}_i + {\bf b}_j)
        \right]
    \end{align}
    Next, firstly, assume that $q_0$ is uniform:
    \begin{align*}
        {\bm h}
        &= \frac{1}{n}\sum_{a = 1}^n\sum_{i = 1}^d \partial_i^2 \phi(\bm{W}{\bf x}_a + \bm{b}) * p_{\varepsilon}\\
        &= -\frac{1}{n\sqrt{M}}\sum_{a = 1}^n\sum_{i = 1}^d \bm{W}_{:, i}^2 \odot \cos(\bm{W}{\bf x}_a + \bm{b}) * p_{\varepsilon}\\
        &= -\frac{1}{n\sqrt{M}}\sum_{a = 1}^n {\rm diag}(\bm{W}\bm{W}^{\top})\odot e^{-\frac{\sigma^2}{2}{\rm diag}(\bm{W}\bm{W}^{\top})} \odot \cos(\bm{W}{\bf x}_a + \bm{b})
    \end{align*}
    For a multivariate normal $q_0(\bf x) = {\cal N}({\bm \mu}, {\bm \Sigma})$, $\nabla\log q_0(\bf x) = -{\bm \Sigma}^{-1}({\bf x} - {\bm \mu})$ there will be additional term to ${\bm h}$
    \begin{align*}
        {\bm h}
        &= \frac{1}{n}\sum_{a = 1}^n\sum_{i = 1}^d \partial_i \phi(\bm{W}{\bf x}_a + \bm{b})\partial_i \log q_0({\bf x}_a) * p_{\varepsilon}\\
        &= -\frac{1}{n\sqrt{M}}\sum_{a = 1}^n\sum_{i = 1}^d \bm{W}_{:, i}\sin(\bm{W}{\bf x}_a + \bm{b})\partial_i \log q_0({\bf x}_a) * p_{\varepsilon}\\
        &= -\frac{1}{n\sqrt{M}} \sum_{a = 1}^n \sin(\bm{W}{\bf x}_a + \bm{b}) \odot \bm{W} \nabla\log q_0({\bf x}_a) * p_{\varepsilon}
    \end{align*}
    using
    \begin{align*}
    \mathbf{w}^\top {\bm \Sigma}^{-1}(\mathbf{x} - {\bm \mu}) \sin (\mathbf{w}^\top \mathbf{x}) * p_{\varepsilon}
    &=
    e^{-\frac{\sigma^2 \|\mathbf{w}\|^2}{2}}
    \mathbf{w}^\top{\bm \Sigma}^{-1}\left [
        (\mathbf{x} - {\bm \mu})\sin(\mathbf{w}^\top \mathbf{x}) + \sigma^2 \mathbf{w}\cos(\mathbf{w}^\top \mathbf{x})
    \right ]
    \end{align*}
    we obtain
    \begin{align}
        {\bm h} = \frac{1}{n\sqrt{M}}e^{-\frac{\sigma^2}{2}{\rm diag}(\bm{W}\bm{W}^{\top})} \odot \sum_{a = 1}^n \left[\sin(\bm{W}{\bf x}_a + \bm{b}) \odot {\bm W}{\bm \Sigma}^{-1}({\bf x}_a - {\bm \mu})
        \right . \nonumber\\
        \left .
        + \sigma^2\cos(\bm{W}{\bf x}_a + \bm{b})\odot {\rm diag}(\bm{W}{\bm \Sigma}^{-1}\bm{W}^{\top})
        \right]
    \end{align}
    In the case of arbitrary $q_0$ we use Taylor expansion:
    \begin{equation*}
        \nabla \log q_0(\mathbf{x} + {\bm \varepsilon}) \approx
        \nabla \log q_0(\mathbf{x}) + \nabla^2 \log q_0(\mathbf{x}) {\bm \varepsilon}
    \end{equation*}
    In the vicinity of $\mathbf{x}$ it is equivalent to previous case and the additional term is obtained with simple replacement: $-{\bm \Sigma}^{-1} \to \nabla^2 \log q_0(\mathbf{x})$ and $-{\bm \Sigma}^{-1}({\bf x} - {\bm \mu}) \to \nabla \log q_0(\mathbf{x})$.

\subsection{Derivation of $\bm H$ and  $\bm h$ \eqref{eq:h} for arc-cosine kernel \eqref{eq:arccos_ker}}
\label{sec:Hh_arccos}
\begin{align*}
        {\bm H}
        &= \partial \Phi^{\top}\partial \Phi * p_{\varepsilon}\\
        &= \sum\limits_{a = 1}^n\sum\limits_{i = 1}^d \partial_i \phi(\bm{W}{\bf x}_a)\partial_i \phi^{\top}(\bm{W}{\bf x}_a) * p_{\varepsilon}\\
        &= \sum\limits_{a = 1}^n\sum\limits_{i = 1}^d \bm{W}_{:, i}\bm{W}_{:, i}^\top \odot {\bf 1}(\bm{W}{\bf x}_a) \odot p^2(\bm{W}{\bf x}_a)^{p - 1}\left((\bm{W}{\bf x}_a)^{p - 1}\right)^{\top} * p_{\varepsilon}\\
        \label{eq:H_derivation}
    \end{align*}
    Considering $p = 2$, uniform base density $q_0$ and isotropic Gaussian noise we will obtain:
    \begin{align*}
        \bm{W}{\bf x}_a{\bf x}_a^{\top}\bm{W}^{\top} * p_{\varepsilon}
        & = \bm{W}\mathbb{E}_{p_{\varepsilon}}({\bf x}_a + {\bm \varepsilon})({\bf x}_a + {\bm \varepsilon})^{\top}\bm{W}^{\top} * p_{\varepsilon} = \bm{W}({\bf x}_a{\bf x}_a^{\top} + \sigma^2{\bf I})\bm{W}^{\top}
    \end{align*}
    The same holds for any symmetric noise distribution with covariance ${\bm \Sigma}$ and corresponding substitution to the above equation.
    \begin{equation}
        {\bm H} = 4\bm{W}\bm{W}^{\top} \odot \sum_{a = 1}^n{\bf 1}(\bm{W}{\bf x}_a) \odot \bm{W}({\bf x}_a{\bf x}_a^{\top} + \sigma^2{\bf I})\bm{W}^{\top}
    \end{equation}
    Moving to the computation of ${\bm h}$ we have:
    \begin{align*}
        {\bm h}
        &= \frac{1}{n}\sum_{a = 1}^n\sum_{i = 1}^d \partial_i^2 \phi(\bm{W}{\bf x}_a) * p_{\varepsilon}\\
        &= \frac{2}{n}\sum_{a = 1}^n\sum_{i = 1}^d \bm{W}_{:, i}^2 \odot {\bf 1}(\bm{W}{\bf x}_a) \odot (\bm{W}{\bf x}_a) * p_{\varepsilon}\\
        &= \frac{2}{n}{\rm diag}(\bm{W}\bm{W}^{\top}) \odot \sum_{a = 1}^n {\bf 1}(\bm{W}{\bf x}_a) \odot (\bm{W}{\bf x}_a)
    \end{align*}
    where the last holds for any symmetric zero-mean density.

%\newpage
\subsection{Taylor approximation of denoising score-matching}
\label{sec:Taylor}
    Considering data corrupted with a small Gaussian noise $\hat{\mathbf{x}} = \mathbf{x} + {\bm \varepsilon}$, ${\bm \varepsilon}\sim {\cal N}(\mathbf{0}, \sigma^2 \mathbf{I})$ and via applying Taylor expansion to the model density $p_m$ we obtain
    %, $s_m(x, \theta) = \nabla_x\log p_m(x, \theta)$
    %\begin{align*}
    %    J_{\varepsilon}(\theta) = \EE_{p_d}\EE_{\varepsilon}\left[{\rm tr}(\nabla_x s_m(\hat{x}, \theta)) + \frac{1}{2}\|s_m(\hat{x}, \theta)\|_2^2\right]
    %\end{align*}
    %\begin{align*}
    %    p_m(x + \varepsilon, \theta) = p_m(x, \theta) + \nabla_x p_m(x, \theta)^{\top} \varepsilon + \frac{1}{2}\varepsilon^{\top}\nabla^2_xp_m(x, \theta)\varepsilon + O(\|\varepsilon\|^3)
    %\end{align*}
    \begin{align*}
        \log p_m(\mathbf{x} + {\bm \varepsilon}, {\bm \theta}) = \log p_m(\mathbf{x}, {\bm \theta}) + \nabla\log p_m(\mathbf{x}, {\bm \theta})^{\top}{\bm \varepsilon} + \frac{1}{2}{\bm \varepsilon}^{\top}\nabla^2\log p_m(\mathbf{x}, {\bm \theta})^{\top}{\bm \varepsilon} + O(\|{\bm \varepsilon}\|_2^3)
    \end{align*}
    where ${\bm \theta}$ denotes a vector of model parameters, $\mathbb{E}[\varepsilon] = {\bf 0}$, $\mathbb{E}[\varepsilon\varepsilon^{\top}] = \sigma^2 {\bf I}$.
    %Assume $\nabla_x^i \EE [\cdots] = \EE \nabla^i_x (\cdots)$, $i = 1, 2$.
    \begin{align*}
        \mathbb{E}_{{\bm \varepsilon}}[\Delta_x\log p_m(\mathbf{x} + {\bm \varepsilon}, {\bm \theta})]
        &\approx \Delta_x \log p_m(\mathbf{x}, {\bm \theta}) + \frac{\sigma^2}{2}\Delta^2_x \log p_m(\mathbf{x}, {\bm \theta})
    \end{align*}
    \begin{align*}
        \|\nabla_x\log p_m(\mathbf{x} + {\bm \varepsilon}, {\bm \theta})\|_2^2
        &=\|\nabla_x\log p_m(\mathbf{x}, {\bm \theta})\|_2^2
        + 2\nabla_x\log p_m(\mathbf{x}, {\bm \theta})^{\top}\nabla_x^2\log p_m(\mathbf{x}, {\bm \theta}){\bm \varepsilon}\\
        &+ {\bm \varepsilon}^{\top}\nabla_x^2\log p_m(\mathbf{x}, {\bm \theta})^{\top}\nabla_x^2\log p_m(\mathbf{x}, {\bm \theta}){\bm \varepsilon}\\
        &+ \nabla_x\log p_m(\mathbf{x}, {\bm \theta})^{\top}\nabla_x{\bm \varepsilon}^{\top}\nabla_x^2\log p_m(\mathbf{x}, {\bm \theta}){\bm \varepsilon} + O(\|{\bm \varepsilon}\|_2^3)
    \end{align*}
    \begin{align*}
        \mathbb{E}_{{\bm \varepsilon}} \|\nabla_x\log p_m(\mathbf{x} + {\bm \varepsilon}, {\bm \theta})\|_2^2
        &\approx \|\nabla_x\log p_m(\mathbf{x}, {\bm \theta})\|_2^2 + \sigma^2{\rm tr}\left[\nabla_x^2\log p_m(\mathbf{x}, {\bm \theta})^{\top}\nabla_x^2log p_m(\mathbf{x}, {\bm \theta})\right] \\
        &+ \sigma^2 \nabla_x\log p_m(\mathbf{x}, {\bm \theta})^{\top}\nabla_x\Delta_x\log p_m(\mathbf{x}, {\bm \theta})
    \end{align*}
    Finally, we have
    \begin{align*}
        J_{{\bm \varepsilon}}({\bm \theta})
        &= J({\bm \theta}) + \frac{\sigma^2}{2}\mathbb{E}_{p_0}\left[(\Delta_x)^2 \log p_m(\mathbf{x}, {\bm \theta})\right]
       + \mathbb{E}_{p_0}{\rm tr}\left[\nabla_x^2\log p_m(\mathbf{x}, {\bm \theta})^{\top}\nabla_x^2\log p_m(\mathbf{x}, {\bm \theta})\right] \\
        &+ \mathbb{E}_{p_0}\left[\nabla_x\log p_m(\mathbf{x}, {\bm \theta})^{\top}\nabla_x\Delta_x\log p_m(\mathbf{x}, {\bm \theta})\right]
    \end{align*}
    where $p_0$ corresponds to an unknown data distribution.
    %$\nabla\log g(x + \varepsilon) = -\Sigma^{-1}(x + \varepsilon - \mu)$
    %\begin{align*}
    %    \nabla\log p(x + \varepsilon)^{\top}\nabla\log g(x + \varepsilon)
    %    \approx -(\nabla\log p(x) + \frac{\gamma}{2}\nabla\Delta \log p(x))^{\top}\Sigma^{-1}(x - \mu) - \gamma{\rm tr}\nabla^2\log p(x)\Sigma^{-1}
    %\end{align*}
    %Let $H = \left[\nabla_xs_m(x, \theta) - s_m(x, \theta)s_m(x, \theta)^{\top}\right]$
    %\begin{align*}
    %    J_{\varepsilon}(\theta) &\approx J(\theta) + \frac{\gamma}{2}\EE_{p_d}\left[\Delta_x{\rm tr}H + {\rm tr}H^{\top}H\right]
    %\end{align*}
%\newpage
\subsection{Nystr\"{o}m kernel approximation}
\label{sec:Nystrom}
Let ${\bm K}$ be a sample Gram matrix, then for Nystr\"{o}m kernel approximation \cite{Chen2016ErrorAO} we have:
\[
    {\bm K} =
    \begin{bmatrix}
        {\bm K}_{11} & {\bm K}_{12}\\
        {\bm K}_{12}^{\top} & {\bm K}_{22}
    \end{bmatrix}
    \quad
    {\bm K} \approx
    \begin{bmatrix}
        {\bm K}_{11}\\
        {\bm K}_{12}^{\top}
    \end{bmatrix}
    {\bm K}_{11}^{-1}
    \begin{bmatrix}
        {\bm K}_{11} & {\bm K}_{12}
    \end{bmatrix}
    \quad
    \phi(x) =
    {\bm K}_{11}^{-\frac{1}{2}}{\bm k}({\bf x})
\]
where ${\bm k}({\bf x}) = \begin{bmatrix} {\bm k}({\bf x}, {\bf x}_1) & \ldots & {\bm k}({\bf x}, {\bf x}_M)\end{bmatrix}^{\top}$, $M$ is the amount of subsampled points.
\[
    \partial {\bm K} = \begin{bmatrix}
        \partial_1 {\bm k}^\top({\bf x}_1) \\
        \cdots \\
        \partial_d {\bm k}^\top({\bf x}_1) \\
        \partial_1 {\bm k}^\top({\bf x}_2) \\
        \cdots \\
        \partial_d {\bm k}^\top({\bf x}_N)
    \end{bmatrix},
    \quad
    \partial \Phi = \partial {\bm K} {\bm K}_{11}^{-\frac{1}{2}}
    \quad
    \partial^2 {\bm K} = \begin{bmatrix}
        \partial_1^2 {\bm k}^\top({\bf x}_1) \\
        \cdots \\
        \partial_d^2 {\bm k}^\top({\bf x}_1) \\
        \partial_1^2 {\bm k}^\top({\bf x}_2) \\
        \cdots \\
        \partial_d^2 {\bm k}^\top({\bf x}_N)
    \end{bmatrix},
    \quad
    \partial^2 {\bm \Phi} = \partial^2 {\bm K} {\bm K}_{11}^{-\frac{1}{2}}
\]
\[
    {\bm G} = \partial {\bm K}^{\top}\partial {\bm K} * p_{\varepsilon},\quad
    {\bm g} = \frac{1}{n}(\partial^2 {\bm K} * p_{\varepsilon})^{\top}{\bf 1}
\]
\begin{align*}
    f
    &= \frac{{\bm k}^{\top}(\cdot)}{\lambda}K_{11}^{-\frac{1}{2}}
    \left[
        {\bm K}_{11}^{-\frac{1}{2}}{\bm g} + ({\bm K}_{11}^{-\frac{1}{2}}{\bm G}{\bm K}_{11}^{-\frac{1}{2}} + n\lambda {\bf I})^{-1}{\bm K}_{11}^{-\frac{1}{2}}G {\bm K}_{11}^{-1}{\bm g}
    \right]\\
    &= \frac{{\bm k}^{\top}(\cdot)}{\lambda}
    \left[
        {\bm K}_{11}^{-1}{\bm g} + ({\bm G} + n\lambda {\bm K}_{11})^{-1}{\bm G} {\bm K}_{11}^{-1}{\bm g}
    \right]\\
\end{align*}

%\section{Old solution VS new one}

%From \cite{GANinstability} and \cite{Hellinger}, $V = \EE_{p_{\varepsilon}}\|\varepsilon\|^2$
%\begin{align*}
%    W(p, q)
%    &\leq V^{\frac{1}{2}} + C\delta(p * p_{\varepsilon}, q * p_{\varepsilon})\\
%    &\leq V^{\frac{1}{2}} + CH(p * p_{\varepsilon}, q * p_{\varepsilon})\\
%    &\leq V^{\frac{1}{2}} + \tilde{C}\sqrt{J(p * p_{\varepsilon}, q * p_{\varepsilon})}\\
%\end{align*}
%\cite{Gretton2013} Fisher divergence minimization implies Hellinger distance minimization, then we need to deconvolve. For gaussian noise $V = n\sigma^2$.

\section{Tables}
\label{sec:B}

\begin{table}[!h]
    \centering
    \begin{tabular}{lcccccccc}
\toprule
Distribution &  \multicolumn{2}{c}{Cosine} & \multicolumn{2}{c}{Uniform} & \multicolumn{2}{c}{Banana}& \multicolumn{2}{c}{Funnel}\\
%Distribution       &          Cosine &     Cosine &         Uniform &    Uniform &          Banana &     Banana &          Funnel &     Funnel &           Rings &      Rings &           Rings &      Rings &         Mixture &    Mixture \\
Model        &  KDSM &  RFFSM &  KDSM &  RFFSM &  KDSM &  RFFSM &  KDSM &  RFFSM \\
\midrule
F$_{train}$&           \textbf{2.197} &      5.331 &           \textbf{1.365} &      1.785 &           0.301 &       \textbf{0.28} &            0.34 &      \textbf{0.288} \\
F$_{test}$  &           \textbf{1.858} &      5.102 &           \textbf{1.584} &      1.901 &           \textbf{0.291} &      0.319 &           0.339 &      \textbf{0.307} \\
LL$_{train}$        &           -5.53 &     -5.008 &           -3.66 &     -3.649 &          -3.529 &     -3.528 &          -2.867 &     -2.846 \\
LL$_{p~train}$    &          -3.528 &     -3.528 &          -3.584 &     -3.584 &           -2.83 &      -2.83 &          -2.868 &     -2.868 \\
LL$_{test}$        &          -5.648 &     -5.056 &          -3.689 &     -3.692 &          -3.659 &     -3.697 &          -2.821 &     -2.783 \\
LL$_{p~test}$     &          -3.503 &     -3.503 &          -3.584 &     -3.584 &          -2.894 &     -2.894 &          -2.796 &     -2.796 \\
FSSD             &          -0.128 &      0.059 &           0.212 &      0.189 &          -0.085 &     -0.045 &           0.058 &     -0.041 \\
p-value          &           \textbf{0.425} &      0.308 &           0.093 &      \textbf{0.131} &           \textbf{0.604} &      0.452 &           0.299 &      \textbf{0.392} \\
W$_1$               &        \textbf{0.251} &   0.372 &       0.06 &  \textbf{0.055} &       \textbf{0.047} &  0.052 &       \textbf{0.06} &  0.084 \\
\bottomrule
\end{tabular}
    \caption{Results of score-matching algorithms, $100$ features and $1000$ sample size for cosine, uniform, banana and funnel distributions.}
    \label{tab:tab:2d_1000_100_1}
\end{table}

\begin{table}[!h]
    \centering
    \begin{tabular}{lcccccc}
     \toprule
     Distribution & \multicolumn{2}{c}{Ring} & \multicolumn{2}{c}{Rings} & \multicolumn{2}{c}{Uniforms}\\ 
     Model &  KDSM &  RFFSM &  KDSM &  RFFSM &  KDSM &  RFFSM \\
     \midrule
     F$_{train}$ &           0.862 &      \textbf{0.635} &           3.664 &      \textbf{3.528} &           \textbf{3.705} &       4.97 \\
    F$_{test}$  &           0.803 &     \textbf{ 0.562} &           3.298 &      3.293 &           \textbf{3.582} &       4.82 \\
    LL$_{train}$         &           -2.35 &     -2.328 &          -3.668 &     \textbf{-4.221} &          \textbf{-3.046} &    -27.879 \\
    LL$_{p~train}$    &          -3.949 &     -3.949 &           -4.68 &      -4.68 &           -2.89 &      -2.89 \\
    LL$_{test}$          &          -2.338 &     -2.346 &          -3.591 &      \textbf{-4.13} &           \textbf{-3.08} &    -27.904 \\
    LL$_{p~test}$     &          -3.929 &     -3.929 &          -4.633 &     -4.633 &           -2.89 &      -2.89 \\
    FSSD             &          -1.316 &     -1.219 &          -0.759 &     -0.851 &           0.057 &     -0.367 \\
    p-value          &           \textbf{0.775} &      0.694 &           \textbf{0.985} &      0.859 &           \textbf{0.347} &      0.673 \\
    W$_1$               &       \textbf{0.063} &  0.086 &        0.212 &   \textbf{0.15} &        \textbf{0.26} &   0.327 \\
     
     \bottomrule
\end{tabular}
    \caption{Results of score-matching algorithms, $100$ features and $1000$ sample size for ring, mixture of rings and mixture of uniforms.}
    \label{tab:2d_1000_100_1}
\end{table}

\section{Figures}
\label{sec:C}
\vspace*{-3in}
% \begin{figure}[!h]
%     \centering
%     \begin{subfigure}[b]{\textwidth}
%       \includegraphics[width=\textwidth]{images/2D/Cosine3500.png}
%       \caption{Cosine}
%     \end{subfigure}

%     \begin{subfigure}[b]{\textwidth}
%       \includegraphics[width=\textwidth]{images/2D/Rings3500.png}
%       \caption{Mixture of rings}
%     \end{subfigure}

%     \caption{Score-matching density estimation using $3500$ sample size for cosine and the mixture of rings deistributions}
%     \label{fig:2d_3500}
% \end{figure}

\begin{figure}[!ht]
    \centering
    \begin{subfigure}[b]{0.32\textwidth}
        \includegraphics[width=\textwidth]{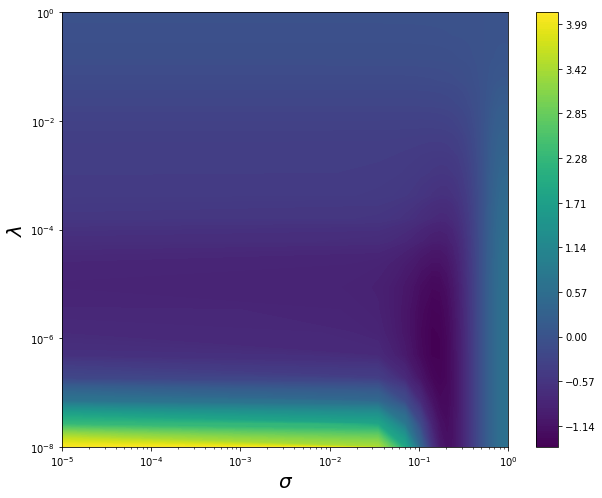}
        \caption{Cosine}
    \end{subfigure}
    \begin{subfigure}[b]{0.32\textwidth}
        \includegraphics[width=\textwidth]{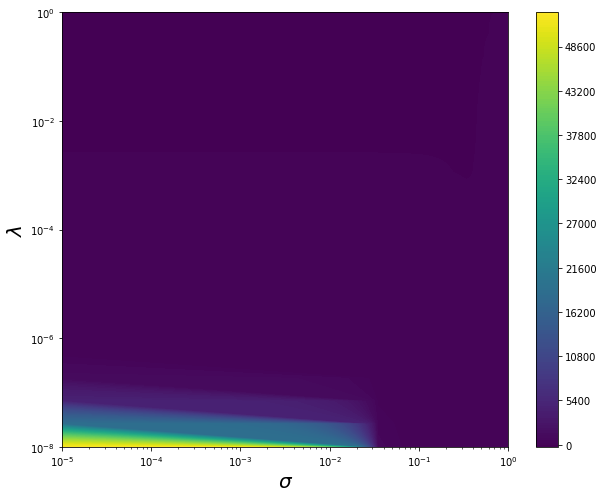}
        \caption{Banana}
    \end{subfigure}
    \begin{subfigure}[b]{0.32\textwidth}
        \includegraphics[width=\textwidth]{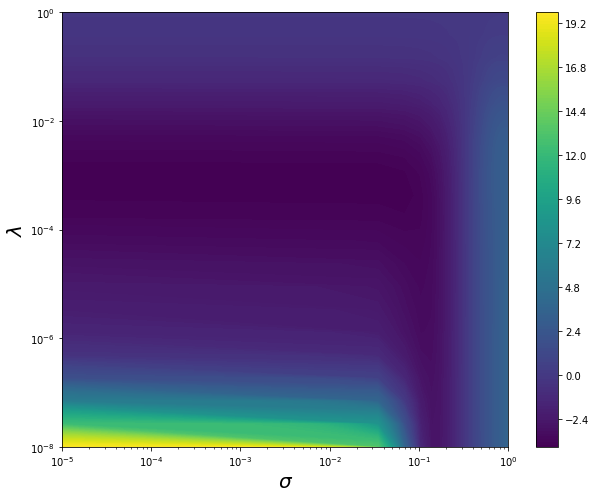}
        \caption{Ring}
    \end{subfigure}
  \begin{subfigure}[b]{0.32\textwidth}
    \includegraphics[width=\textwidth]{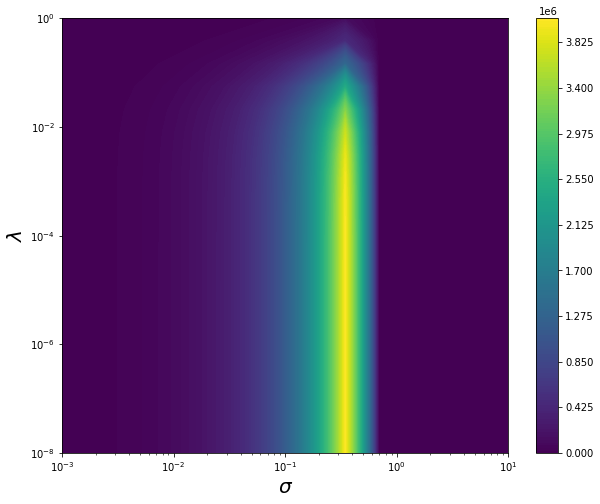}
    \caption{MiniBoone}
  \end{subfigure}
    \begin{subfigure}[b]{0.32\textwidth}
        \includegraphics[width=\textwidth]{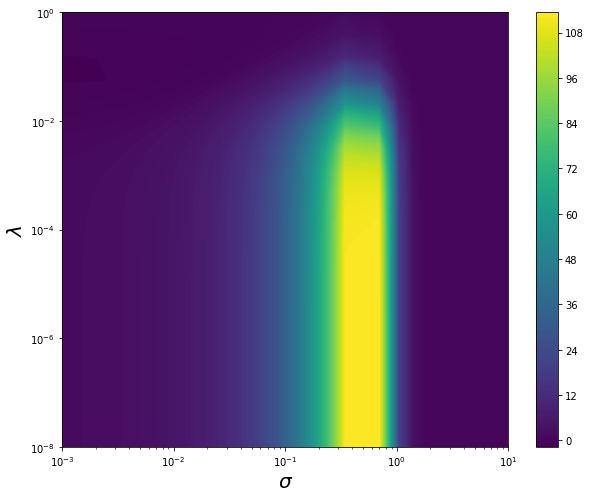}
        \caption{Red Wine}
    \end{subfigure}
    \begin{subfigure}[b]{0.32\textwidth}
        \includegraphics[width=\textwidth]{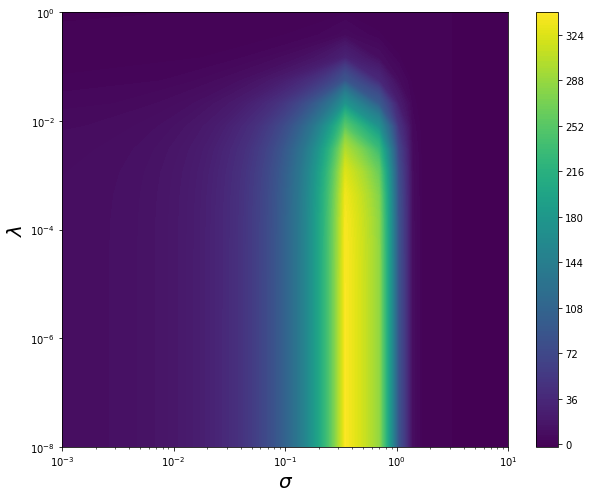}
        \caption{White Wine}
    \end{subfigure}
    \caption{Loss surface w.r.t. regularization parameter $\lambda$ (y axis) and noise parameter $\sigma$ (x axis).}
    \label{fig:lossdemo_app}
\end{figure}

\begin{figure}[!ht]
    \centering
    \begin{subfigure}[b]{0.8\textwidth}
        \includegraphics[width=\textwidth]{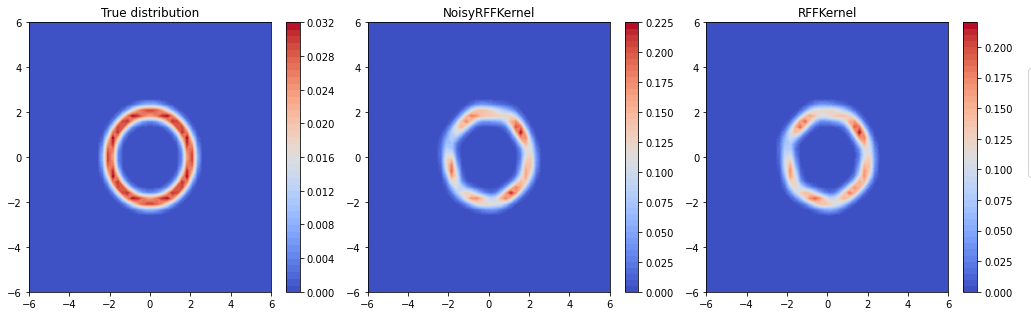}
        \caption{Ring}
    \end{subfigure}

    \begin{subfigure}[b]{0.8\textwidth}
        \includegraphics[width=\textwidth]{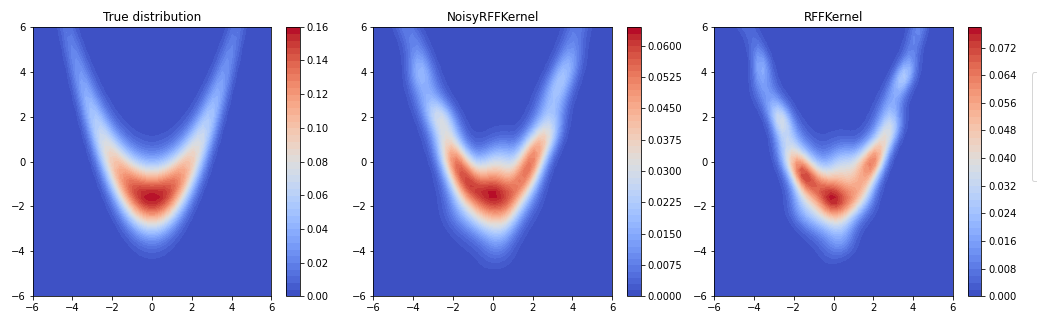}
        \caption{Banana}
    \end{subfigure}

    \begin{subfigure}[b]{0.8\textwidth}
        \includegraphics[width=\textwidth]{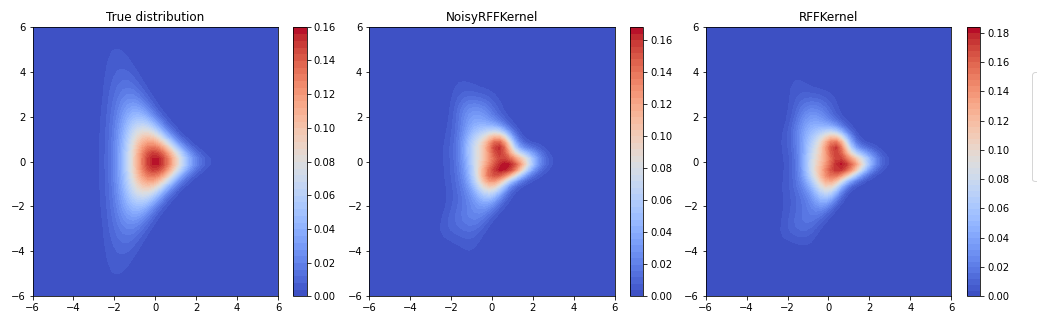}
        \caption{Funnel}
    \end{subfigure}

    \begin{subfigure}[b]{0.8\textwidth}
        \includegraphics[width=\textwidth]{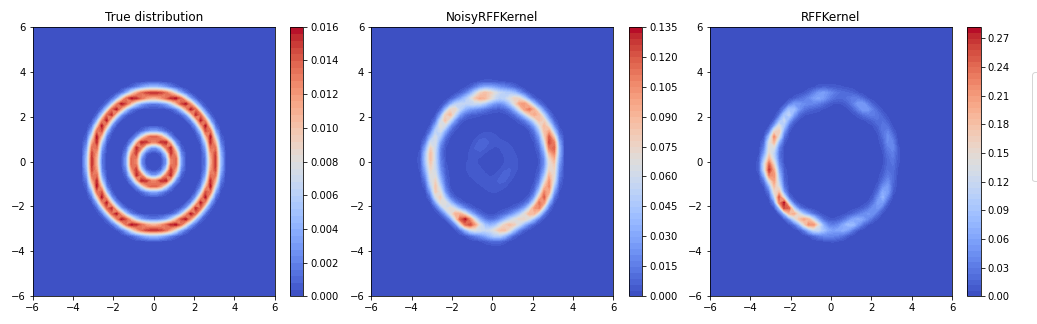}
        \caption{Mixture of rings}
    \end{subfigure}

    \begin{subfigure}[b]{0.8\textwidth}
        \includegraphics[width=\textwidth]{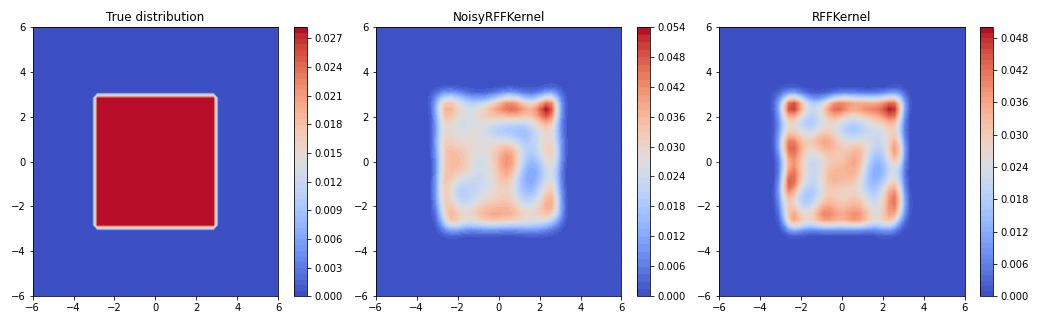}
        \caption{Uniform}
    \end{subfigure}

    \caption{Score-matching density estimation using $1000$ samples.
    Left column is a ground truth, middle is DSM RFf,
    right~--- SM RFF.}
    \label{fig:2d_1000_app}
\end{figure}
\end{document}